\documentclass{article}
\usepackage{amssymb}
\usepackage{times}
\usepackage{graphicx}
\usepackage{enumitem}       
\usepackage{csquotes}       
\usepackage{amsfonts}       
\usepackage{nicefrac}       
\usepackage{amsmath}        
\usepackage{bbm}
\usepackage{mathrsfs}
\usepackage{gensymb}
\usepackage{physics}
\usepackage{url}            
\usepackage{wrapfig}        
\usepackage{multicol}
\usepackage{multirow}
\usepackage{makecell}
\newcommand\mydata[2]{$#1_{\pm#2}$}
\usepackage[table,xcdraw]{xcolor} 
\definecolor{ColorA}{HTML}{e5f4fc}
\definecolor{ColorB}{HTML}{f4e3f4}
\usepackage{caption}
\usepackage[final]{corl_2025} 

\newcommand{\eps}{\epsilon}

\newcommand{\E}{\mathbb{E}}
\newcommand{\R}{\mathbb{R}}

\newcommand{\obj}{O}

\newcommand{\pose}{\vb*{p}}
\newcommand{\initcand}{\Tilde{\pose}}
\newcommand{\cand}{\hat{\pose}}

\newcommand{\data}{\mathcal{D}}
\newcommand{\dist}{p_{\text{data}}}
\newcommand{\energy}{E_{\theta}}
\newcommand{\score}{\vb*{\Phi}_{\theta}}

\newcommand{\loss}{\mathcal{L}}

\newcommand{\estprior}{\pi_{\text{est}}}
\newcommand{\trackprior}{\pi_{\text{track}}}

\newcommand{\g}{\vb*{g}}

\def\ie{\emph{i.e}.}

\newcommand\jiyao[1]{\textcolor{black}{}} 
\newcommand\rebuttal[1]{\textcolor{black}{#1}}

\newcommand\nnfootnote[1]{%
  \begin{NoHyper}
  \renewcommand\thefootnote{}\footnote{#1}%
  \addtocounter{footnote}{-1}%
  \end{NoHyper}
}

\DeclareMathOperator*{\argmax}{arg\,max}
\DeclareMathOperator*{\argmin}{arg\,min}

\title{UniTac2Pose: A Unified Approach Learned in Simulation for Category-level Visuotactile 
  In-hand Pose Estimation}

\author{
    Mingdong Wu\textsuperscript{ \rm 1,2*}, 
    Long Yang\textsuperscript{ \rm 1,2*}, 
    Jin Liu\textsuperscript{ \rm 3,4}, 
    Weiyao Huang\textsuperscript{\rm 1,2}, \\
    \textbf{Lehong Wu}\textsuperscript{\rm 1,2},
    \textbf{Zelin Chen}\textsuperscript{\rm 3,4}, 
    \textbf{Daolin Ma}\textsuperscript{\rm 3,4}, 
    \textbf{Hao Dong}\textsuperscript{\rm  1,2 $\dagger$}
    \\
  $^1$ Center on Frontiers of Computing Studies, School of Computer Science, Peking University \\
  $^2$ PKU-AgiBot Lab, 
  $^3$ School of Ocean and Civil Engineering, Shanghai Jiao Tong University, \\
  $^4$ Xense Robotics
}

\begin{document}

\maketitle

\vspace{-20pt}


\begin{abstract}
Accurate estimation of the in-hand pose of an object based on its CAD model is crucial in both industrial applications and everyday tasks, ranging from positioning workpieces and assembling components to seamlessly inserting devices like USB connectors. While existing methods often rely on regression, feature matching, or registration techniques, achieving high precision and generalizability to unseen CAD models remains a significant challenge. In this paper, we propose a novel three-stage framework for in-hand pose estimation. The first stage involves sampling and pre-ranking pose candidates, followed by iterative refinement of these candidates in the second stage. In the final stage, post-ranking is applied to identify the most likely pose candidates. These stages are governed by a unified energy-based diffusion model, which is trained solely on simulated data. This energy model simultaneously generates gradients to refine pose estimates and produces an energy scalar that quantifies the quality of the pose estimates. Additionally, borrowing the idea from the computer vision domain, we incorporate a render-compare architecture within the energy-based score network to significantly enhance sim-to-real performance, as demonstrated by our ablation studies. We conduct comprehensive experiments to show that our method outperforms conventional baselines based on regression, matching, and registration techniques, while also exhibiting strong intra-category generalization to previously unseen CAD models. Moreover, our approach integrates tactile object pose estimation, pose tracking, and uncertainty estimation into a unified framework, enabling robust performance across a variety of real-world conditions. 
\nnfootnote{*: equal contribution, $\dagger$: corresponding author}
\end{abstract}

\vspace{-5pt}


\vspace{-10pt}
\section{Introduction}
\vspace{-5pt}

Precise pick-and-place operations are essential in industrial applications and daily tasks, from positioning workpieces and assembling components to inserting devices like USB connectors. This requires the robot to localize the objects' in-hand pose based on its CAD model with high precision~\cite{bauza2023tac2pose}. A promising approach estimates the in-hand pose using high-resolution tactile sensors, which provide rich geometrical contact information and are highly discriminative~\cite{bauza2023tac2pose}. Unlike vision-based methods~\cite{wen2024foundationpose}, tactile-based approaches are immune to extrinsic calibration errors or occlusions.

Robotic in-hand object pose estimation is a challenging task due to several factors: the need to handle previously unseen objects, the requirement to regrasp when the estimated pose has high uncertainty~\cite{bauza2024simple_error}, and the necessity to track pose changes caused by extrinsic collisions with the environment.
A common approach is to train a pose regressor using tactile images and the object’s CAD model on a simulated dataset. However, due to the localized nature of tactile sensing, this often leads to significant ambiguity—multiple poses can result in similar tactile imprints. Classical registration methods such as ICP and FilterReg~\cite{bauza2019tactile, gao2023hand, gao2019filterreg} attempt to align tactile point clouds with the global object model. Yet, these methods are highly sensitive to initialization and prone to local minima. Moreover, tactile-based depth estimates are inherently noisy due to the monocular nature of most sensors, further degrading performance.
Another line of work~\cite{bauza2023tac2pose, gao2023hand, bauza2024simple_error} explores feature-matching strategies. For example, Tac2Pose~\cite{bauza2019tactile} converts tactile RGB images into binary contact maps and matches them to rendered maps from pose grids using features learned via MoCo~\cite{he2020momentum}. However, these maps are noisy, often object-specific, and fail to capture in-plane surface features—such as a triangular ridge on a flat surface, especially when the object is under significant deformation.
\begin{wrapfigure}{tr}{0.50\textwidth}
\begin{center}
\vspace{-0pt}
\includegraphics[width=\linewidth, trim=0cm 0cm 0cm 0cm, clip]{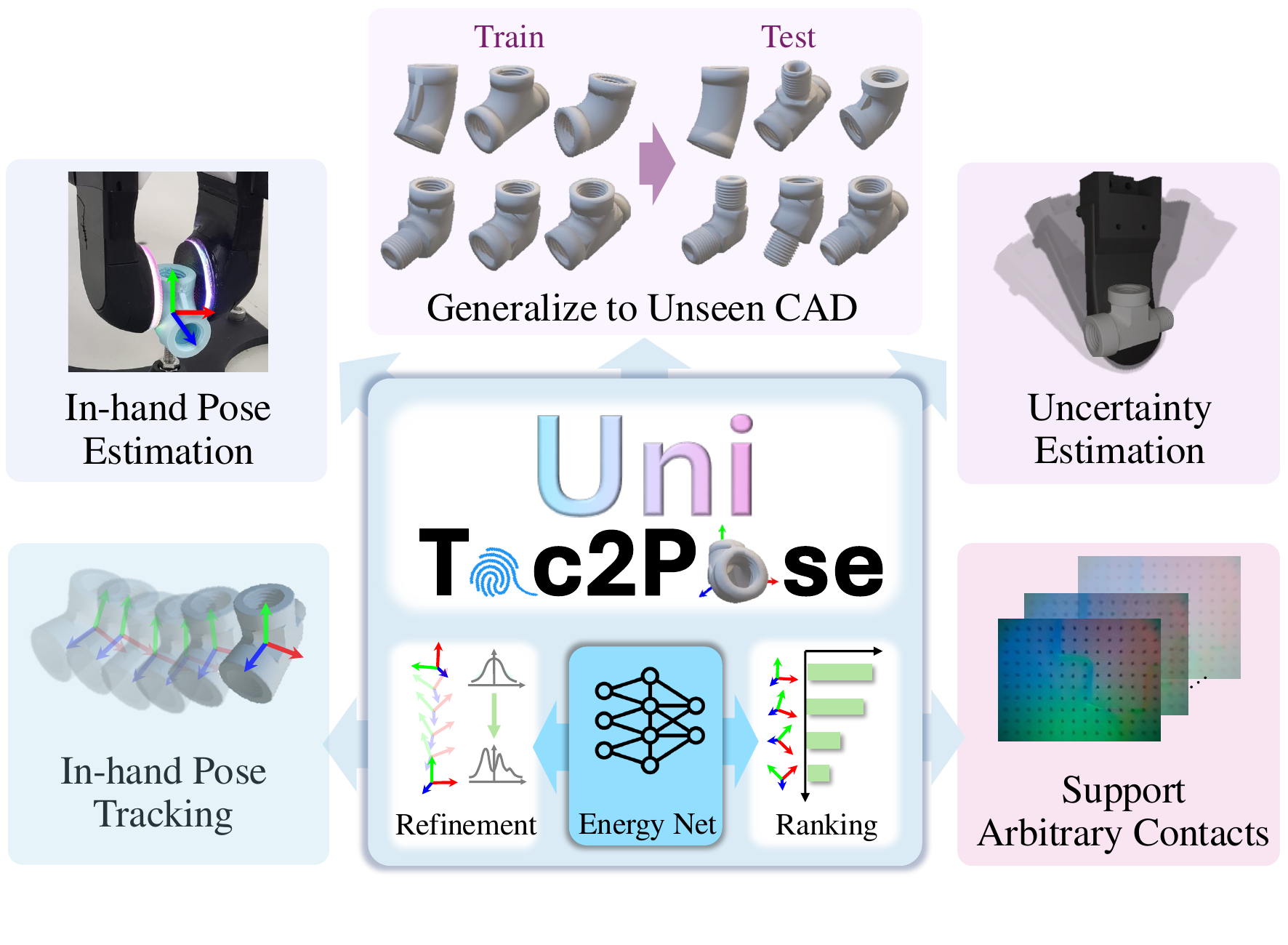}
\end{center}
\vspace{-20pt}
\caption{
The core of UniTac2Pose is an energy-based diffusion model that unifies tactile pose estimation, tracking, and uncertainty, conditioned on multi-contact and generalizable to unseen CADs.
}
\label{fig:teaser}
\vspace{-10pt}
\end{wrapfigure}

We propose UniTac2Pose, a unified framework for in-hand object pose estimation, tracking, and uncertainty estimation, as illustrated in \rebuttal{Figure~\ref{fig:teaser}}. At its core is an energy-based diffusion model that estimates the (unnormalized) log-likelihood of an object pose, conditioned on observed tactile imprints and the object’s CAD model. This scalar energy score enables candidate pose ranking, while its gradient indicates the optimal direction for refinement.
The framework consists of three stages: pre-ranking, refinement, and post-ranking. During inference, we first sample initial pose candidates from a prior distribution, such as the one in~\cite{wen2024foundationpose}, and pre-rank them using the energy network. Next, the top candidates are refined through gradient-based optimization guided by the energy network. Finally, a post-ranking stage selects the most likely pose. The energy network is trained end-to-end using a score-matching objective on purely simulated data. To bridge the sim-to-real gap, we adopt a render-and-compare architecture that significantly improves performance, as validated by our ablation studies.

Our approach offers several key advantages:
\textbf{Unified functionality.} UniTac2Pose integrates pose estimation, tracking, and uncertainty quantification. For example, pose tracking is achieved by centering the initial pose prior on the previous prediction, while pose uncertainty is quantified via the variance of refined pose candidates.
\textbf{End-to-end training.} Unlike prior methods that rely on feature matching or intermediate regressors, our method directly optimizes for pose likelihood using end-to-end training, avoiding compounding errors.
\textbf{Generalization to unseen intra-category objects.} By conditioning on multiple tactile imprints and leveraging simulated training, our framework generalizes effectively to unseen CAD models without requiring real-world data.

We validate our method through extensive real-world experiments. UniTac2Pose consistently outperforms regression-, matching-, and registration-based baselines in both pose estimation and tracking tasks, and demonstrates strong generalization across object categories.

In summary, our contributions are as follows: 
\begin{itemize}[itemsep=5pt,topsep=0pt,parsep=0pt, leftmargin=15pt] 
\item We introduce UniTac2Pose, a unified framework for visuotactile in-hand object pose estimation, tracking, and uncertainty quantification, capable of handling diverse contact scenarios and generalizing to unseen objects. 
\item We propose a novel three-stage framework based on an energy-based diffusion model trained solely on simulated data, requiring no real-world supervision. 
\item Our method achieves state-of-the-art performance in real-world pose estimation and tracking, and to the best of our knowledge, is the first to demonstrate intra-category generalization in visuotactile pose estimation. \end{itemize}

\vspace{-10pt}
\section{Related Works} 
\vspace{-10pt}
\label{sec:related_works}

\subsection{Tactile Object Pose Estimation}
\vspace{-10pt}
Tactile perception has evolved significantly, from early work using low-resolution, often planar or bulky sensors~\cite{schaeffer2003methods,corcoran2010tracking,petrovskaya2011,chalon2013online,Bimbo2016,saund2017touch,javdani2013efficient,chebotar2014learning}, to modern high-resolution tactile sensing for more refined pose estimation. Early approaches often relied on binary contact signals and required repeated readings~\cite{koval2015,Koval2017}, limiting robustness and efficiency. Subsequent methods combined vision and tactile sensing to improve global pose estimation~\cite{bimbo2015global,Allen1999,Ilonen2014,Falco2017,yu2018realtime}. However, in many of these, tactile input served merely as a binary signal to refine visual estimates, inherently bounding accuracy to the visual modality. In contrast, recent efforts focus on using high-resolution tactile data as a primary modality, enabling contact shape recovery~\cite{platt2011using,pezzementi2011,Luo2017}, or leveraging deformable tactile surfaces for improved localization~\cite{Kuppuswamy2019FastMC}. Distributed tactile arrays have also been explored to enhance in-hand object tracking~\cite{tu2023posefusion,li2023vihope}.

Our work builds on GelSlim~\cite{donlon2018gelslim}, a high-resolution visual-tactile sensor previously applied in grasp assessment~\cite{Hogan2018}, 3D shape reconstruction~\cite{Wang2018}, and tactile-based manipulation~\cite{tian2019manipulation}. Registration-based pose estimation methods using tactile point clouds~\cite{bauza2019tactile,yang2023hand} have shown promise, but remain sensitive to initialization and are affected by depth noise inherent in tactile data.
The most relevant prior line of work is Tac2Pose and its variants~\cite{villalonga2021tactile,bauza2023tac2pose,bauza2024simple_error}, which formulate pose estimation as a probabilistic inference problem. Our method differs fundamentally by learning a unified energy-based model in an end-to-end fashion, avoiding the need to chain pose refinement steps and enabling category-level generalization to unseen object instances via CAD model conditioning.

\vspace{-10pt}
\subsection{Applications of Score-based Generative Models}
\vspace{-10pt}
Score-based generative models have emerged as powerful tools for modeling complex data distributions via gradient estimation of log-likelihoods~\cite{hyvarinen2005estimation,denosingScoreMatching}. Denoising score matching (DSM)\cite{denosingScoreMatching} was introduced to improve training stability, followed by sliced score matching\cite{song2020sliced}, annealed training~\cite{song2019generative}, and other training refinements~\cite{song2020improved}. These advances culminated in diffusion-based models that perform particularly well in high-dimensional domains such as image synthesis~\cite{song2020score}, leveraging continuous-time stochastic processes for generation.

More recent efforts have broadened the applicability of these models to 3D and robotics tasks, including point cloud generation~\cite{cai2020learning}, denoising~\cite{luo2021score}, depth completion~\cite{shao2022diffustereo}, and human pose estimation~\cite{ci2022gfpose}. Notably, works such as GenPose~\cite{zhang2024generative} have demonstrated how score-based and energy-based models can be integrated for robust 6D pose estimation under uncertainty, using diffusion sampling guided by learned energy fields.
In contrast, our approach trains a single, unified energy network that simultaneously computes gradients for iterative pose refinement and outputs scalar energy values to assess pose quality. Additionally, we condition this network on object CAD models, enabling render-and-compare strategies for instance-specific generalization—a key difference from category-level-only approaches such as GenPose.

\begin{figure*}[t]
\begin{center}
\includegraphics[width=0.9\linewidth, trim=0cm 0cm 0cm 0cm,  clip]{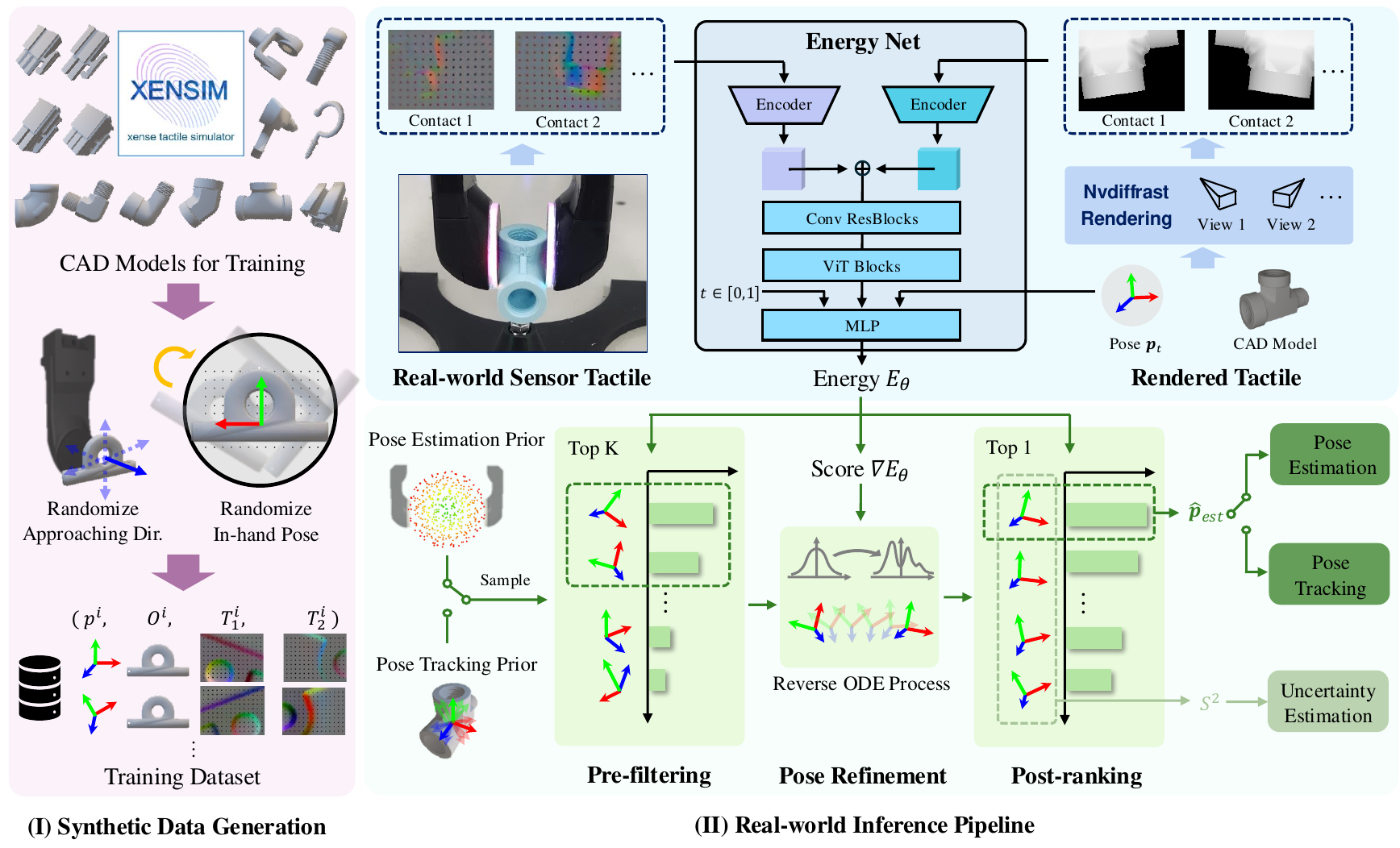}
\vspace{-12pt}
\end{center}

\caption{\textbf{Method Overview.} (I): We first generate a synthetic dataset using the FEM-based tactile simulator XENSIM. We randomly sample in-hand poses to generate a diverse training dataset with pure simulation. (II): During inference, the Energy Net takes real-world tactile, rendered tactile, object pose and diffusion timestep as inputs, and outputs the energy and score of the pose. For pose estimation and tracking, we sample N pose candidates from a prior distribution, and get the final pose by pre-filtering, refinement and post-ranking. For uncertainty estimation, we calculate the variance of refined poses to represent the uncertainty of the grasp.\jiyao{Visualize the render-compared refinement process, (t=xxx, t=aaa, ... )}}
\label{fig:pipeline}
\vspace{-15pt}
\end{figure*}

\vspace{-10pt}
\section{Method} 
\label{sec:method}
\vspace{-10pt}

\textbf{Problem Statement:}
This work addresses the task of estimating the 6D pose of in-hand objects within the Tool Center Point (TCP) frame, crucial for robotic pick-and-place operations. Given a CAD model of an object $O$, we aim to estimate the 6D in-hand pose $\pose \in \R^{3+6}$ from tactile imprints $T = (T_1, T_2, \dots, T_k)$, where $k$ is the number of tactile contacts. We represent the pose as a 9-D variable $[R | T]$ following~\cite{zhang2024generative}, with $R \in \R^6$ for rotation and $T \in \R^3$ for translation. To handle discontinuities in quaternion and Euler angle representations, we use a continuous 6-D rotation representation $[R_x | R_y]$~\cite{chen2021fs, zhou2019continuity}.

\textbf{Overview:}
We first generate a synthetic dataset using the FEM-based tactile simulator XENSIM, represented as $\data = \{ (\pose_i, \obj_i, T_i)\}_{i=1}^n$, where $\pose_i \in \text{SE}(3)$ denotes the 6D pose, and $\obj_i \in \R^{3 \times N}$ is the canonical object point cloud. The tactile observations $T \in \R^{3 \times H \times k \times W}$ are concatenated tactile images from GelSlim 3.0~\cite{donlon2018gelslim}, with $H$ and $W$ representing the image height and width.

In Sec.~\ref{sec:abcde}, we describe the training of an energy-based diffusion model $\energy$ on the synthetic dataset. During inference (Sec.~\ref{sec:inference}), given a CAD model $\obj^*$ and tactile imprints $T^*$, we estimate the 6D pose in three stages. First, initial pose candidates are sampled from a prior distribution~\cite{wen2024foundationpose}, and low-likelihood candidates are discarded, leaving ${ \cand_1, \cand_2, ..., \cand_K }$, with $K=16$. These candidates are then refined using gradients from the energy network $\energy$. Finally, the candidates are ranked by energy, and the one with the lowest energy is selected as the final pose. Additionally, our framework also supports pose tracking and uncertainty estimation~(Appendix~\ref{appendix:extension}).

\vspace{-10pt}
\subsection{Synthesizing Tactile Pose Estimation Dataset}
\label{sec:dataset}
\vspace{-10pt}

We synthesize the tactile pose estimation dataset $\data = \{ (\pose_i, \obj_i, T_i)\}_{i=1}^n$ for each training object via simulating the RGB tactile image under different object-to-hand poses~(\ie, grasp poses) in a FEM-based tactile simulation. Due to the page limit, we defer data generation details to Appendix~\ref {appendix:sim_dataset}.

\vspace{-10pt}
\subsection{Training the Energy-based Diffusion Model}
\label{sec:abcde}
\vspace{-10pt}

UniTac2Pose is a three-stage framework unified by an energy-based diffusion model $\energy$, trained on the synthesized dataset $\data = \{ (\pose_i, \obj_i, T_i)\}_{i=1}^n$ as described above. 
The energy network $\energy: \R^{3+6} \times \R^{3\times N} \times \R^{3 \times H \times k \times W} \times [0, 1] \to \R^1$ takes as input an intermediate pose variable $\pose$, the canonical object point cloud $\obj$, the concatenated tactile images $T$ and a continuous time variable $t \in [0, 1]$, and outputs a scalar energy value $\energy(\pose, \obj, T, t) \in \R^1$. 

\textbf{Training Objective:}
We assume the synthetic dataset $\data$ is sampled from an implicit data distribution $\data = \{(\pose_i, \obj_i, T_i) \sim \dist(\pose, \obj, T)\}$.
The energy network is trained to match between the derivative of its output $\nabla_{\pose} \log \energy(\pose, \obj, T, t)$ regarding the input pose variable and the \textit{score function} of the perturbed conditional distribution $\nabla_{\pose} \log p_{t}(\pose|\obj, T)$, for all $t \in [0, 1]$, using the score-matching training objective~\cite{song2019generative,song2020sliced,song2020score}.

The energy network $\energy$ is then trained via the following Denoising Score Matching~(DSM)~\cite{denosingScoreMatching} objective:
\begin{equation}
\begin{aligned}
   &\loss(\theta) = \E_t
   \left\{
   \lambda(t)\E
   \left[ \left\Vert\nabla_{\pose}\energy(\pose(t), t , \obj, T)  - \frac{\pose(0) - \pose(t)}{\sigma(t)^2} \right\Vert_2^2 \right] \right\}
\end{aligned}
\label{eq:score_matching_loss}
\end{equation}
where $\pose(0) \sim \dist(\pose(0)|\obj, T)$, $\pose(t) \sim \mathcal{N}(\pose(t);\pose(0), \sigma^2(t)\mathbf{I})$, and $\eps=10^{-5}$ is a hyper-parameter that denotes the minimal noise level.

\textbf{Architectural Design:}
Instead of directly encoding all inputs using MLPs, CNNs, or PointNet, we introduce a key design, \ie, render-compare, into the energy network $\energy$, inspired by the impressive results of FoundationPose~\cite{wen2024foundationpose}.
During inference, the pose variable $\pose(t)$ is first normalized to an $\text{SE}(3)$ pose $\overline{\pose}(t)$.
Next, in the TCP frame, the canonical CAD model of the object is initialized at $\overline{\pose}(t)$, and RGB images $I = (I_1, I_2, \dots, I_K)$ are rendered from multiple pre-calibrated cameras using Nvdiffrast~\cite{Laine2020diffrast}. 
Note that the camera poses are based on pre-calibrated real sensor cameras.
Unlike FoundationPose, we encode the horizontally concatenated tactile images and rendered images using two separate encoders, as they come from different modalities.
Similar to FoundationPose, the features extracted by these CNNs are concatenated and subsequently fed into Convolutional Residual Blocks and 
\rebuttal{
ViT~\cite{dosovitskiy2020vit} 
}
Blocks.
We adopt a parameterization trick, proven effective for training energy-based diffusion models in~\cite{zhang2024generative, salimans2021should}, as follows:
\begin{equation}
\begin{aligned}
    \energy(\pose, \obj, T, t) = \langle \score(\pose, \obj, T, t), \pose \rangle
\end{aligned}
\label{eq:parameterization}
\end{equation}
where $\score: \R^{3+6} \times \R^{3\times N} \times \R^{3 \times H \times k \times W} \times [0, 1] \to \R^{3+6}$ is the main backbone of the energy network, as described above, and outputs a 9-dimensional vector, which matches the dimension of the pose variable.

\vspace{-10pt}
\subsection{Tactile In-hand Pose Estimation Using the Energy Network}
\label{sec:inference}
\vspace{-10pt}

According to~\cite{denosingScoreMatching}, minimizing the objective in Eq.~\ref{eq:score_matching_loss} leads to an optimal energy network $\energy^*$ that satisfies the following equation under some mild assumptions:
\begin{equation}
\begin{aligned}
    \nabla_{\pose}\energy^*(\pose, \obj, T, t) = \nabla_{\pose} \log  p_{t}(\pose | \obj, T) \ \implies \ \energy^*(\pose, \obj, T, t) = \log  p_{t}(\pose | \obj, T) + C(\obj, T)
\end{aligned}
\label{eq:optimal}
\end{equation}
where $C(\obj, T)$ is a constant independent of $\pose$.
Although the optimal energy model differs from the ground truth likelihood by the constant $C(\obj, T)$, it still serves as a reliable surrogate likelihood estimator for ranking candidates, given fixed tactile observations $T$ and the object CAD model $\obj$:
\begin{equation}
\begin{aligned}
\energy^*(\pose_i, \obj, T, \eps) &> \energy^*(\pose_j, \obj, T, \eps)
\iff   \log  p_{\eps}(\pose_i|\obj, T) &> \log  p_{\eps}(\pose_j|\obj, T)
\end{aligned}
\label{eq:optimal_energy}
\end{equation}
Motivated by these results, we propose a three-stage framework that utilizes the energy network's output for pose ranking in both pre-filtering and post-ranking and the gradient from the energy network for the pose refinement stage.

\textbf{Pre-filtering}
Our framework begins by sampling a large set of initial pose candidates from a prior distribution $\estprior$. 
For pose estimation, we use the prior distribution from~\cite{wen2024foundationpose} for global pose sampling. 
Specifically, we first uniformly sample $N_s$ viewpoints from an icosphere centered on the object, with the camera facing the center, and then convert these camera-to-object poses into object-to-TCP poses.
Next, using the trained energy model, we rank the candidates $\{\initcand_i\}^M_{i=1}$ into a sequence $ \initcand_1 \succ \initcand_2 \dots \succ  \initcand_M$, where:
\begin{equation}
\begin{aligned}
\initcand_i \succ \initcand_j \iff \energy(\initcand_i, \obj, T, \eps) > \energy(\initcand_j, \obj, T, \eps)
\end{aligned}
\label{eq:energy_distillation_loss}
\end{equation}
and $M$ is a hyperparameter.
We then filter out the bottom $M-K$ of candidates, leaving $\{\initcand_i\}^K_{i=1}$ as the output of the pre-filtering stage, where $K$ is another hyperparameter.

\textbf{Pose Refinement}  
Further, we iteratively refine the candidates using gradients derived from the trained energy network. Specifically, we refine $\{\cand_i\}^K_{i=1}$ via a modified version of the \textit{Probability Flow} (PF) ODE~\cite{song2020score}, integrating from $t = t_0$ to $\eps$, to obtain the refined candidates $\{\cand_i\}^K_{i=1}$:  
\begin{equation}
\begin{aligned}
    \frac{d\pose}{dt} = - \sigma(t)\dot{\sigma}(t)\nabla_{\pose}\log p_{t}(\pose | \obj)
\end{aligned}
\label{eq:reverse_sde}
\end{equation}
Here, the ODE starts from the initial pose candidates $\{\initcand_i\}^K_{i=1}$, where $t_0$ is a hyperparameter. 
The score function $\nabla_{\pose}\log p_{t}(\pose | \obj)$ is approximated by the gradient of the energy network $\energy(\pose, \obj, T, t)$, and the ODE is solved using the RK45 solver~\cite{dormand1980family}.  
Since the initial candidates are already close to the high-density regions of $\dist(\pose| \obj, T)$, we start the refinement from a smaller $t$, where the gradient is more informative.  

\textbf{Post-Ranking}  
Finally, we rank the refined pose candidates $\{\cand_i\}^K_{i=1}$ using the energy network, similar to the pre-filtering stage. 
Unlike GenPose~\cite{zhang2024generative}, which aggregates predictions from multiple candidates, we select the single candidate with the highest energy as the final pose estimate:  
\begin{equation}
\begin{aligned}
\cand_{est} = \argmax_{i} \energy(\cand_i, \obj, T, \eps)
\end{aligned}
\label{eq:post_ranking}
\end{equation}

\subsection{Extending the Framework for Pose Tracking and Uncertainty Estimation}
Our framework can be extended to support pose tracking and uncertainty estimation, enhancing object manipulation robustness. Due to the page limit, we defer the complete derivations and discussions to Appendix~\ref{appendix:extension}. The key ideas are as follows:

\textbf{Pose Tracking.}
We track the in-hand object pose by modifying the prior distribution $\trackprior(\initcand_{t+1})$ in the pre-filtering stage, incorporating temporal continuity from previous poses. This method enables real-time tracking at 10 Hz with fewer refinement steps.

\textbf{Uncertainty Estimation.}
Uncertainty is estimated by computing the variance $S^2$ of the pose candidates $\{\cand_i\}^K_{i=1}$. The grasp with the lowest uncertainty is selected for re-grasping. Relative uncertainty between grasps is determined by comparing the $S^2_i$ values, as shown in Eq.~\ref{eq:energy_distillation_loss}.

\section{Results} 
\label{sec:results}

\subsection{Experiment Setup} 
\label{sec:setup}


\textbf{Objects and Synthetic Dataset.}
We generate synthetic and real-world evaluation datasets using 30 objects from the McMaster dataset.  
Following~\cite{villalonga2021tactile, corona2018pose}, all objects are sourced from McMaster.  
We evaluate our method against baselines on 10 distinct objects, including symmetric shapes and those with localized features prone to inducing singularities.  
For each object, we generate 20,000 data points.  
To assess category-level generalization, we include two common industrial assembly categories: \textit{pipe} and \textit{connector}.  
The \textit{pipe} category consists of 13 objects, with 8 for training and 5 for testing.  
The \textit{connector} category consists of 7 objects, with 5 for training and 2 for testing.  
For each training object in \textit{pipe}, we generate 10,000 data points, while for each training object in \textit{connector}, we generate 5,000 data points.  
Each category contains structurally similar objects, allowing the network to learn shared features from the training set and generalize to unseen objects.

\textbf{Real World Evaluation.}  
Following~\cite{bauza2023tac2pose}, we use the GelSlim 3.0 for both simulation and real-world data collection. 
The hardware setup is detailed in Appendix~\ref{appendix:real_world_setup}.
For each object, we first collect a fixed number of grasps, recording both the tactile images and the robot's TCP pose.  
To obtain the ground truth object pose in the TCP frame, we replicate the real-world scene in simulation, manually annotate the first few ground-truth poses, and automatically generate the rest. We defer the detailed procedure to Appendix~\ref{appendix:real_world_setup}.  
In total, we collect over 3800 grasps across all 30 objects.

\rebuttal{Since the objects in our study include symmetric shapes, which introduce ambiguity in pose estimation (as multiple configurations of symmetric objects can appear identical), we use ADD-S as the evaluation metric. ADD-S is an adaptation of the ADD (Average Distance of Model Points) metric. Specifically, we apply ADD-S for objects with symmetry and ADD for objects without symmetry. For simplicity, we refer to this unified metric as ADD-S in our experiments.}
\begin{equation}
    \text{ADD-S} = \frac{1}{|\mathcal{M}|} \sum_{\mathbf{p} \in \mathcal{M}} \min_{\mathbf{q} \in \mathcal{M}} \| (\mathbf{R}_{\text{est}} \mathbf{p} + \mathbf{t}_{\text{est}}) - (\mathbf{R}_{\text{gt}} \mathbf{q} + \mathbf{t}_{\text{gt}}) \|
\label{eq:add-s}
\end{equation}
As shown in Eq.~\ref{eq:add-s}, ADD-S addresses this challenge by considering the closest matching point on the ground-truth object when calculating distances, rather than relying on a one-to-one correspondence.

\textbf{Baselines.}  
Nevertheless, we implement 3 typical approaches from previous studies, \ie, \textit{FilterReg}, \textit{Vanilla Regression}, and \textit{Matching~(Tac2Pose~\cite{bauza2023tac2pose})} with same network architecture and capabilities to ensure a fair comparison. Implementation details are deferred to the Appendix.~\ref{appendix:baseline}.

\def\isMainDocument{}
\ifdefined\isMainDocument
\else
\documentclass{article}

\usepackage{array}
\usepackage{multicol}
\usepackage{multirow}
\usepackage{graphicx}
\usepackage{makecell}

\newcommand\mydata[2]{$#1_{\pm#2}$}
\begin{document}

\fi

\begin{table*}[tbph!]
\centering
\renewcommand{\arraystretch}{1.5} 

\caption{\textbf{Instance-level sim-to-real evaluation results evaluation results.} \rebuttal{We report ADD-S (mm) and ADD (mm) errors for symmetric and non-symmetric objects respectively. Lower ADD/ADD-S error implies better performance.}
\jiyao{clarify ADD for non-symmetric, ADD-S for symmetric objects. Add a new line to clarify the input modality of each method. Reply: this has been added to intro of ADD \& ADD-S on page 7.} We compare our method with \rebuttal{FilterReg, regression, and matching} as baselines. We also conduct ablation studies on our key designs, i.e., using depth as rendered input, unifying score and energy network, render-compare, and RGB augmentation.}

\resizebox{\textwidth}{!}{
\begin{tabular}{|c c|c c | c c c|c c c| c |}
    \hline
    & & \multicolumn{2}{c|}{Ours} & \multicolumn{3}{c|}{Ablations} & \multicolumn{3}{c|}{Baselines} & Oracle \\
    && RGB & Depth & w/o Unify & w/o R-C & w/o Aug  & \rebuttal{FilterReg(Global)} & Regression & \rebuttal{Matching} & \rebuttal{FilterReg(Partial)} \\ \hline

    
    Bear Housing  & \makecell{\begin{minipage}{10mm}\vspace{1pt}\centering\includegraphics[height=7mm]{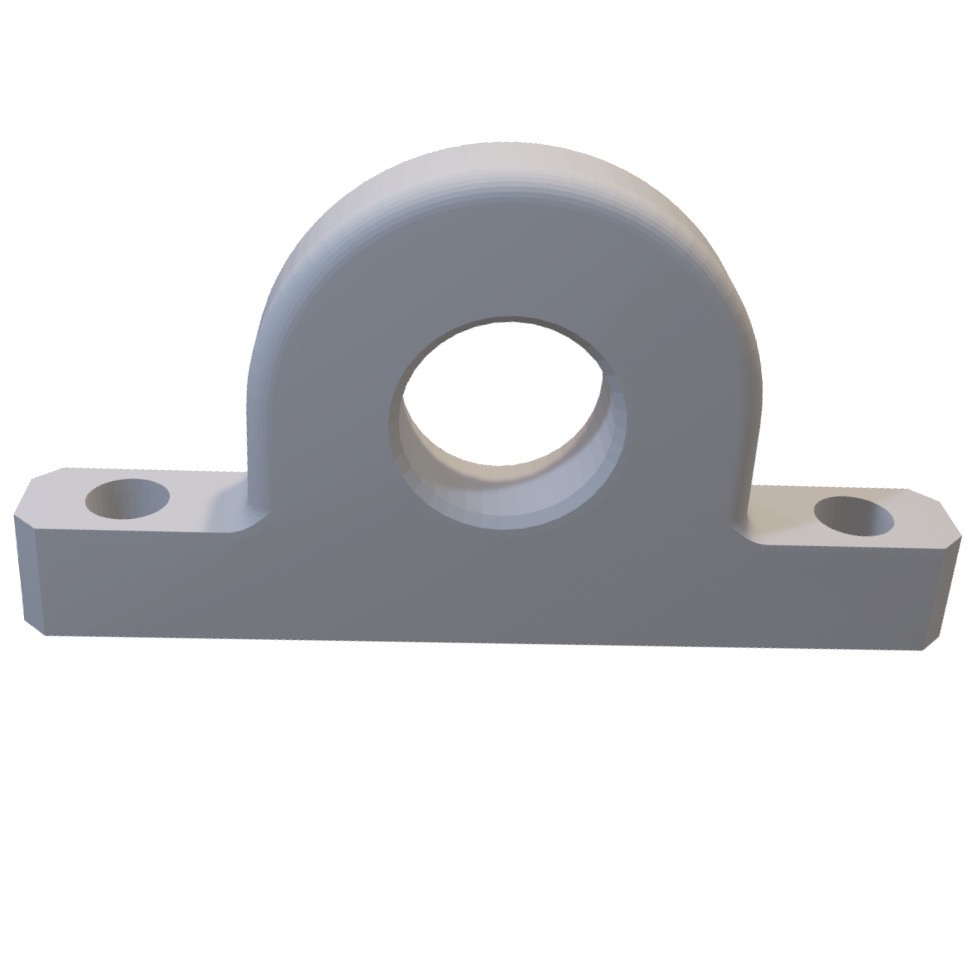}\vspace{1pt}\\\end{minipage}} & \textbf{2.5} & \underline{2.7} & 8.7 & 5.3 & 8.7 & 6.9 & 10.8 & 13.9 & 2.2   \\ \hline
    Cable Holder  & \makecell{\begin{minipage}{10mm}\vspace{1pt}\centering\includegraphics[height=7mm]{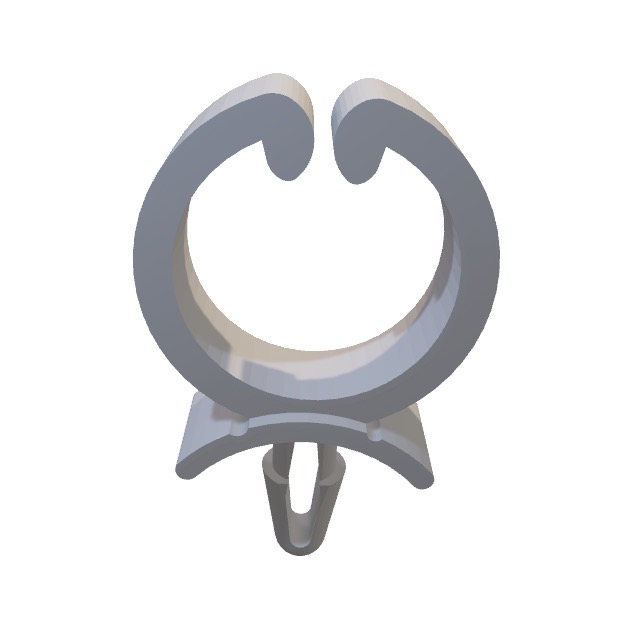}\vspace{1pt}\\\end{minipage}} & \underline{2.7} & \textbf{2.0} & 7.8 & 9.9 & 3.5 & 7.8 & 3.1 & 12.9 & 2.8  \\ \hline
    Cable Clip  & \makecell{\begin{minipage}{10mm}\vspace{1pt}\centering\includegraphics[height=7mm]{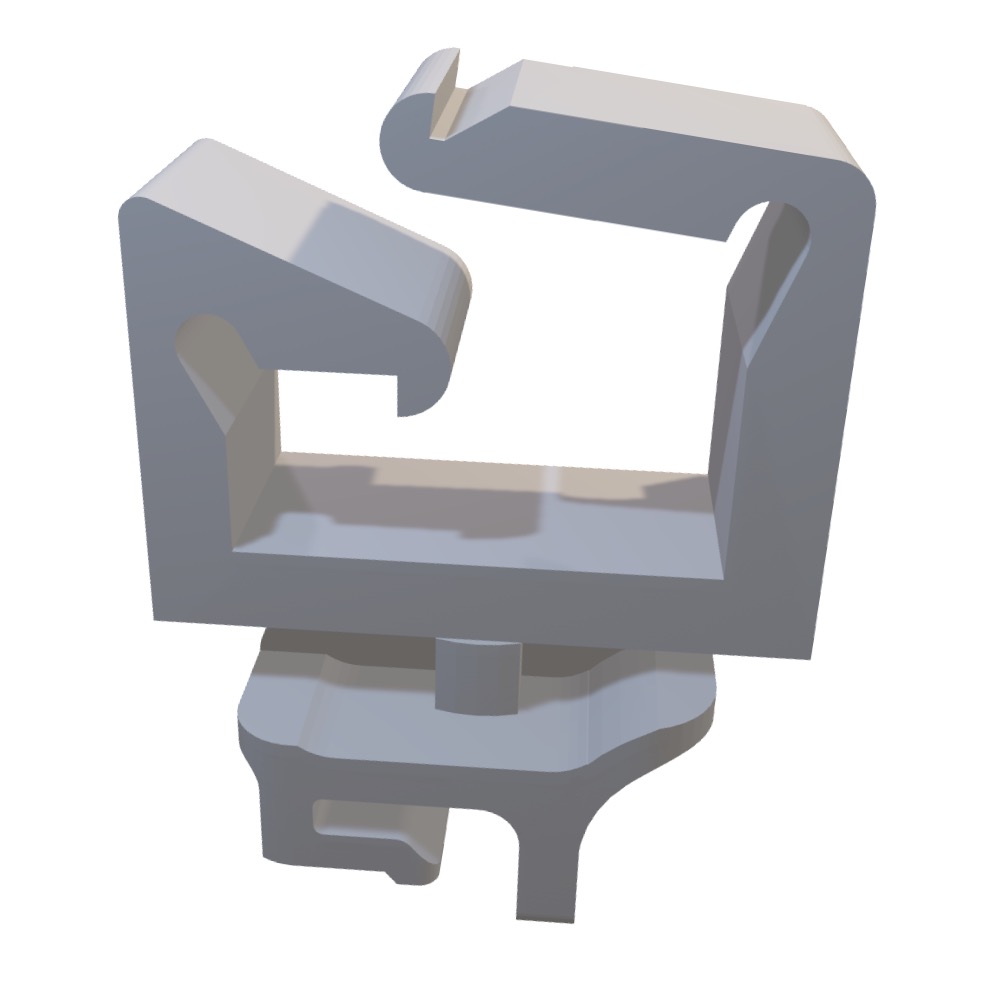}\vspace{1pt}\\\end{minipage}} & \textbf{1.6} & 1.9 & 2.2 & 1.9 & \textbf{1.6} & 8.2 & 10.4 & 12.7 & 2.2 \\ \hline      
    Round Nut  & \makecell{\begin{minipage}{10mm}\vspace{1pt}\centering\includegraphics[height=7mm]{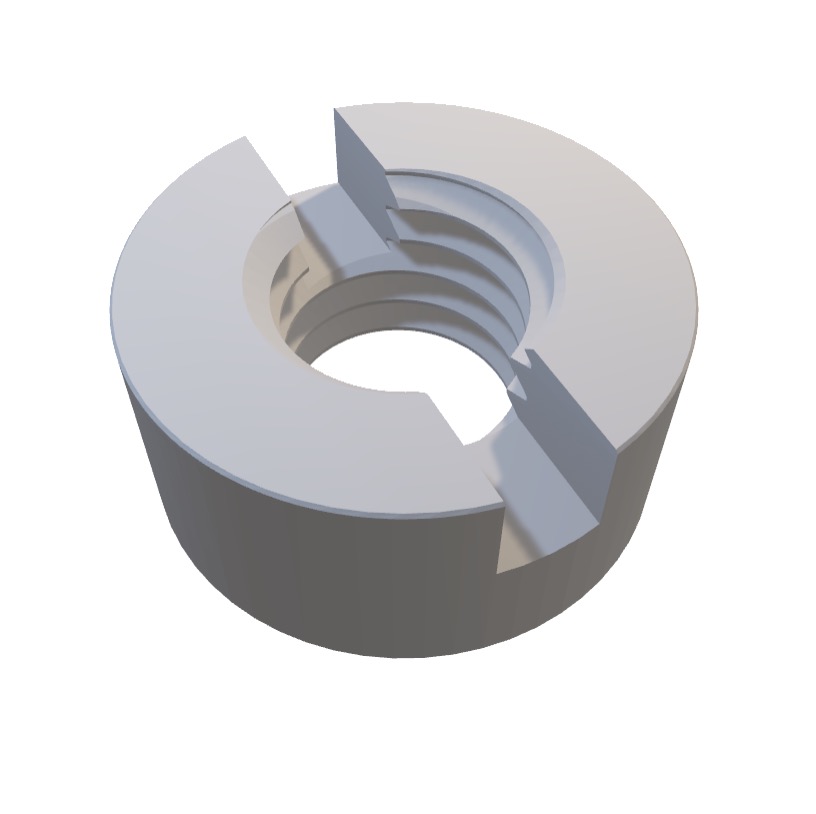}\vspace{1pt}\\\end{minipage}} & \textbf{2.4} & 3.6 & 3.3 & 43.2 & \underline{2.5}  & 5.4 & 45.4 & 8.4 & 2.2  \\ \hline      
    Cotter  & \makecell{\begin{minipage}{10mm}\vspace{1pt}\centering\includegraphics[height=7mm]{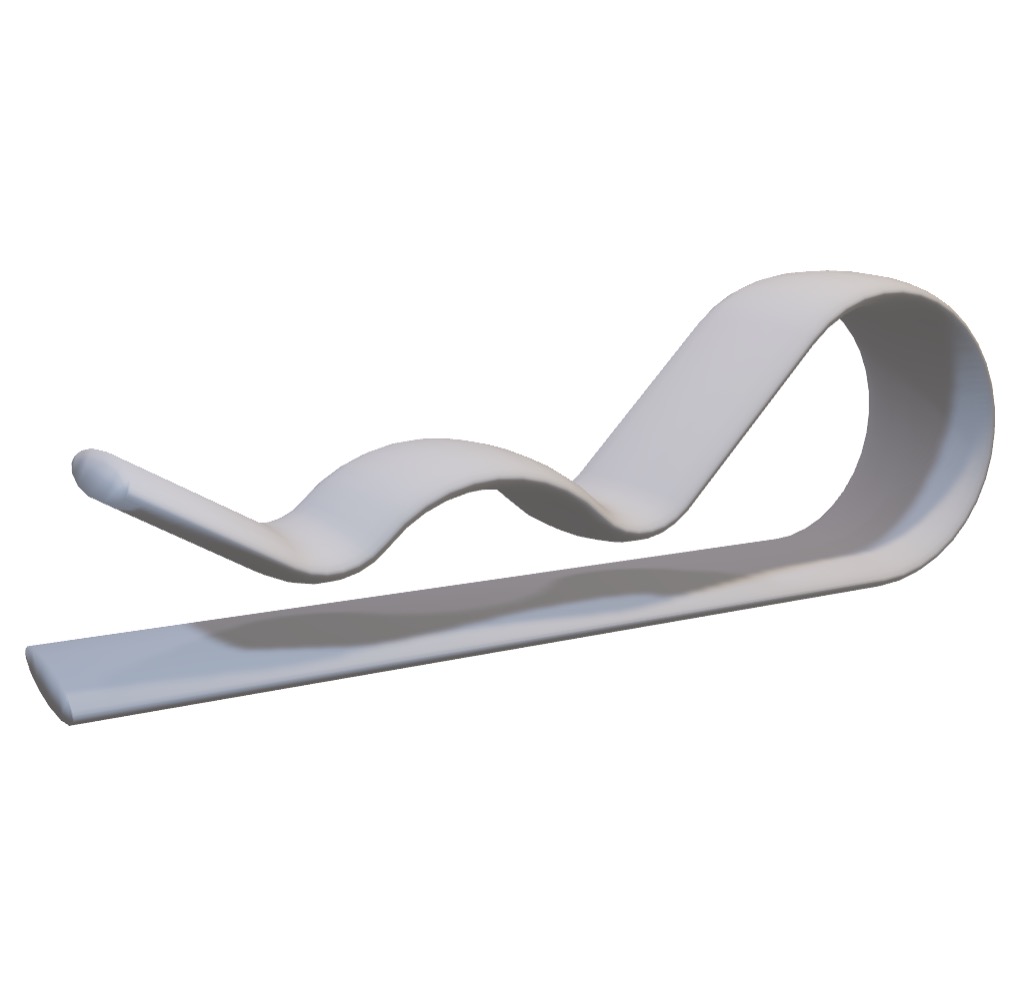}\vspace{1pt}\\\end{minipage}} & \textbf{1.5} & \underline{1.8} & 4.1 & 7.5 & 18.1 & 21.0 & 3.9 & 30.2 & 3.5  \\ \hline      
    Hook  & \makecell{\begin{minipage}{10mm}\vspace{1pt}\centering\includegraphics[height=7mm]{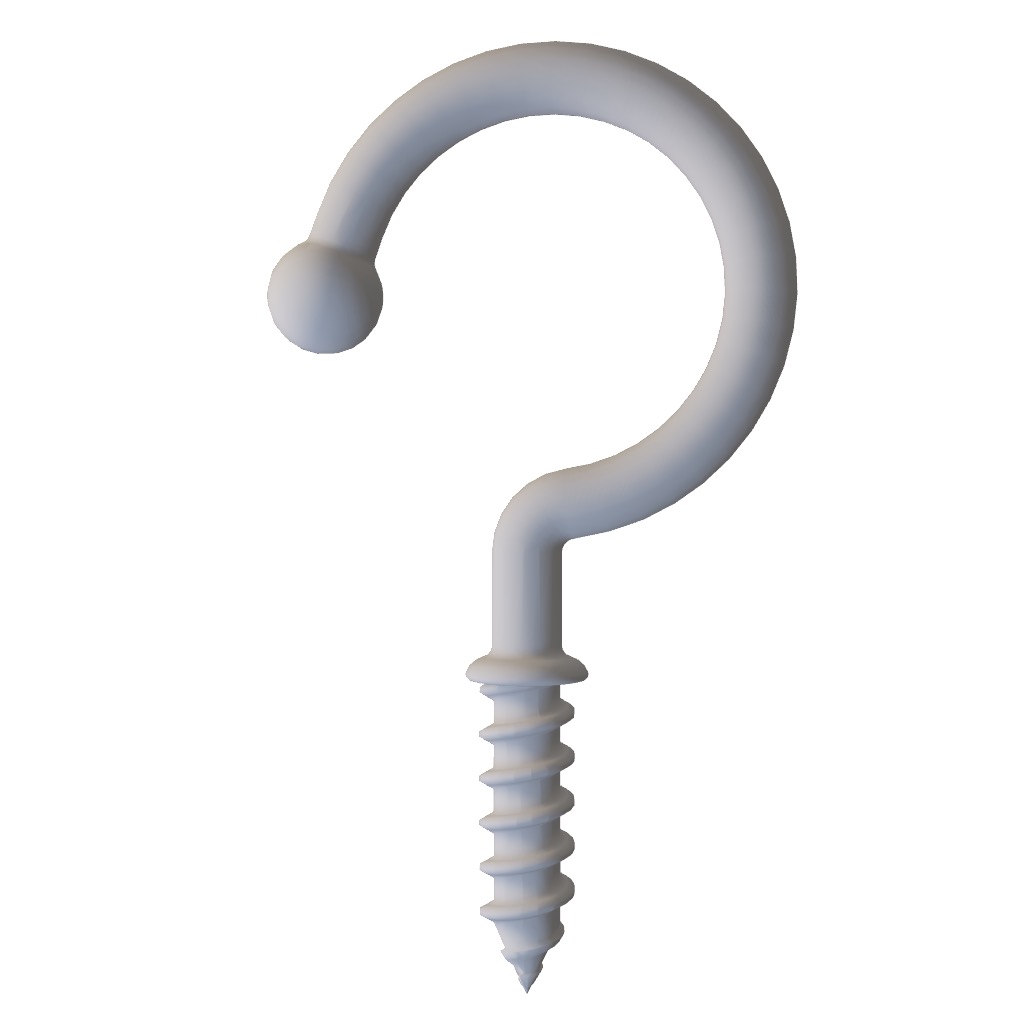}\vspace{1pt}\\\end{minipage}} & \textbf{3.0} & 9.7 & \underline{3.4} & 29.9 & 32.7 & 20.7 & 35.2 & 36.9 & 2.7  \\ \hline      
    Hose  & \makecell{\begin{minipage}{10mm}\vspace{1pt}\centering\includegraphics[height=7mm]{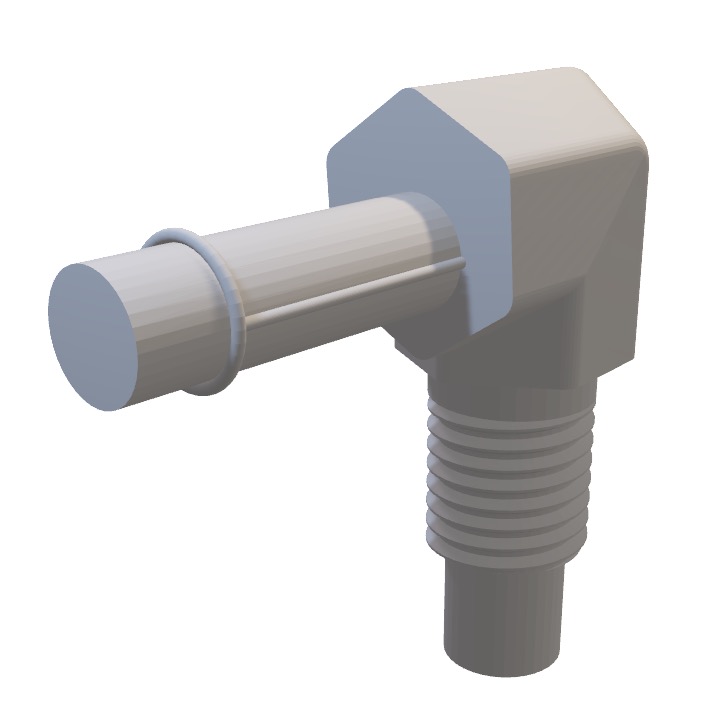}\vspace{1pt}\\\end{minipage}} & 1.5 & \underline{1.2} & \textbf{1.0} & 3.2 & 1.4  & 20.7 & 1.8 & 13.9 & 1.6 \\ \hline      
    Hydraulic  & \makecell{\begin{minipage}{10mm}\vspace{1pt}\centering\includegraphics[height=7mm]{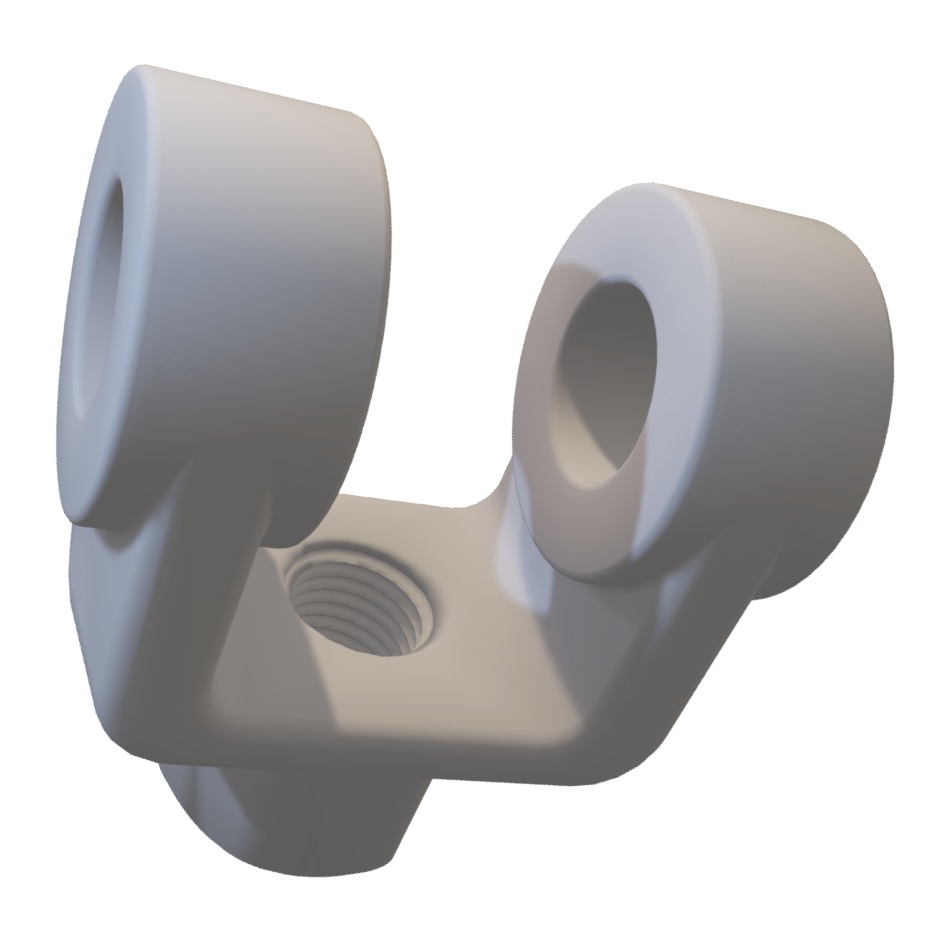}\vspace{1pt}\\\end{minipage}} & \underline{2.4} & \textbf{2.3} & 2.8 & 3.5 & 2.7  & 10.0 & 5.3 & 13.3 & 3.2  \\ \hline      
    Stud  & \makecell{\begin{minipage}{10mm}\vspace{1pt}\centering\includegraphics[height=7mm]{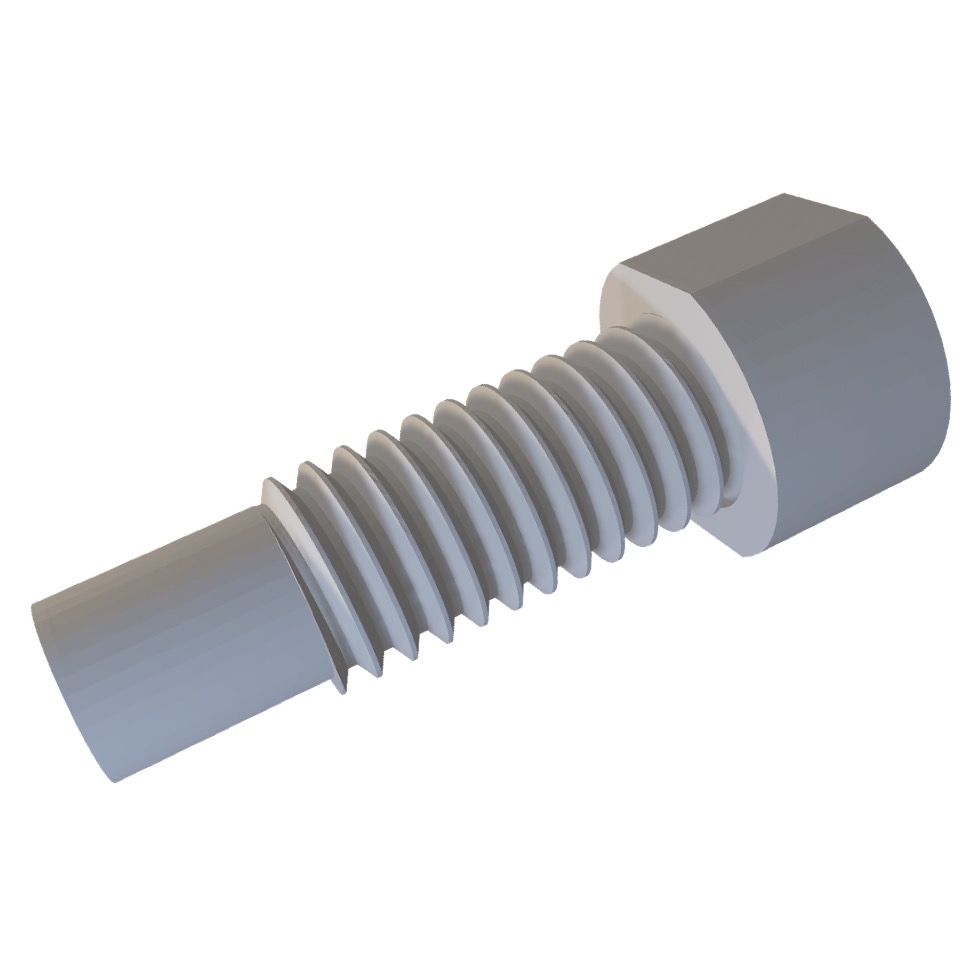}\vspace{1pt}\\\end{minipage}} & \textbf{2.2} & 9.7 & \underline{3.6} & 18.3 & 9.9  & 26.1& 8.0 & 12.4  & 6.9  \\ \hline      
    Rail  & \makecell{\begin{minipage}{10mm}\vspace{1pt}\centering\includegraphics[height=7mm]{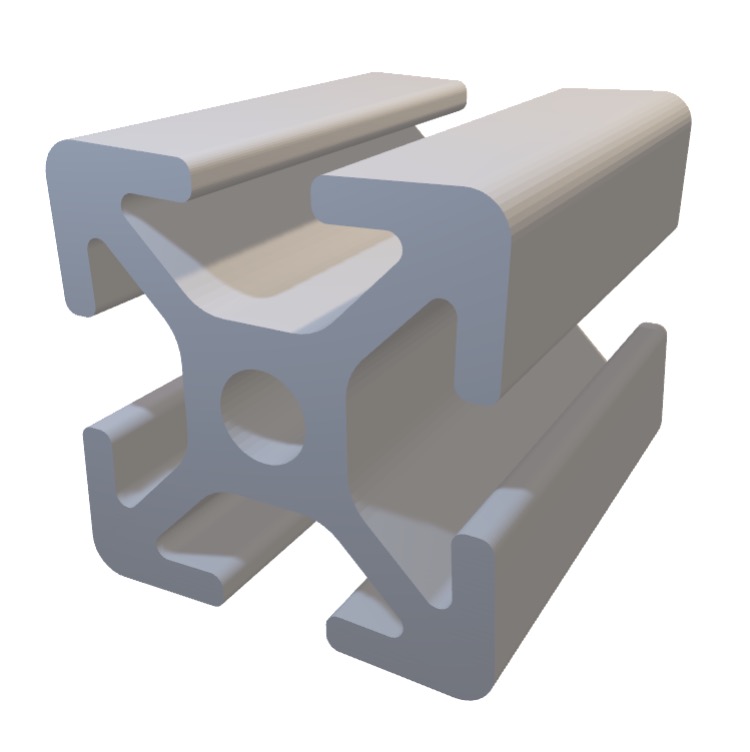}\vspace{1pt}\\\end{minipage}} & \underline{1.5} & 1.8 & \textbf{1.1} & 2.5 & 1.8 & 5.9 & 6.1 & 8.0 & 4.4  \\ \hline 
    Average & & \textbf{2.1} & \underline{3.7} & 3.8 & 12.5 & 8.3 & 13.3 & 13.0 & 16.3 & 3.2\\ \hline
\end{tabular}
}

\label{table:main_results}
\vspace{-10pt}
\end{table*}

\ifdefined\isMainDocument
\else
\end{document}
\fi

\vspace{-5pt}
\subsection{Generalizable Tactile Pose Estimation within a Single Instance}
\vspace{-5pt}
\label{sec:baseline-ablations}

\begin{figure*}[t]
\begin{center}
\vspace{-5pt}
\includegraphics[width=0.9\linewidth, trim=0cm 5cm 0cm 0cm,  clip]{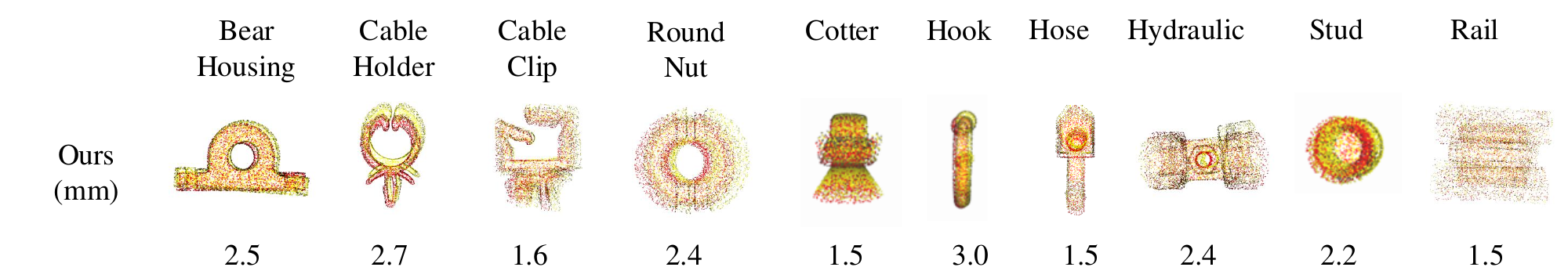}
\end{center}
\vspace{-10pt}
\caption{\textbf{Visualization of ADD errors.} We visualize point clouds of objects with ground truth poses (red) and estimated poses (yellow). The ADD errors are the same as reported in Tab. \ref{table:main_results}.}
\vspace{-15pt}
\label{fig:add_vis}
\end{figure*}

We evaluate UniTac2Pose for tactile pose estimation at the instance level, training on synthesized datasets of a single object and testing on real-world tactile data from the same object, with potentially different grasps. Objects from McMaster, such as the \textit{Cable Clip} and \textit{Cotter}, as well as symmetrical objects like \textit{Bear Housing} and \textit{Round Nut}, are used.

\textbf{Comparison with Baselines.} As shown in Tab.\ref{table:main_results}, our method outperforms all baselines across all objects, demonstrating superior performance in real-world tactile pose estimation. While the \textit{Regression} baseline works well for small objects with distinct features, it struggles with symmetric or larger ones. \textit{FilterReg (Global)} fails on nearly all objects due to global point cloud registration issues, whereas \textit{FilterReg (Partial)} performs comparably to our method. The \textit{Matching} baseline underperforms due to discrepancies between simulation and real-world contact shapes, similar to issues in Tac2Pose\cite{bauza2023tac2pose}. Our approach, powered by score-matching training~\cite{zhang2024generative}, handles these challenges effectively, providing robust predictions even with ambiguous or symmetric tactile data.

\textbf{Ablation Studies.}
We compare our method against several ablations: removing the render-compare mechanism (\textit{Ours w/o R-C}), training a separate score-based diffusion network for pose refinement instead of using the unified approach (\textit{Ours w/o Unify}), and disabling data augmentation (\textit{Ours w/o Aug(mentation)}). As shown in Tab.~\ref{table:main_results}, our method consistently outperforms all of these variants. The most significant performance drop occurs when the render-compare mechanism is removed, highlighting its crucial role. Disabling data augmentation leads to a performance decline, especially for objects like \textit{Cotter} and \textit{Hook}, while other objects are less impacted. The variant without the unified approach (\textit{Ours w/o Unify}) performs similarly to the full model but shows slightly higher ADD-S errors, demonstrating the importance of our unified energy network in effectively aligning the pose refinement, ranking, and evaluation stages.
We also conduct further ablations on the three stages (i.e., pre-filtering, refinement, and post-ranking) and the choice of hyperparameters in Appendix~\ref{appendix:ablation_three_stages} and Appendix~\ref{appendix:ablation_t}, respectively.

\textbf{Robustness of the Render-Compare Modality.}
We evaluate the robustness of two alternative energy network versions that render RGB images and depth maps. Both yield comparable performance, but RGB-based rendering outperforms depth-based rendering for objects like \textit{Stud} and \textit{Hook}, suggesting it is more stable in tactile pose estimation.

\begin{figure*}[t]
\begin{center}
\includegraphics[width=\linewidth, trim=0cm 0cm 0cm 0cm,  clip]{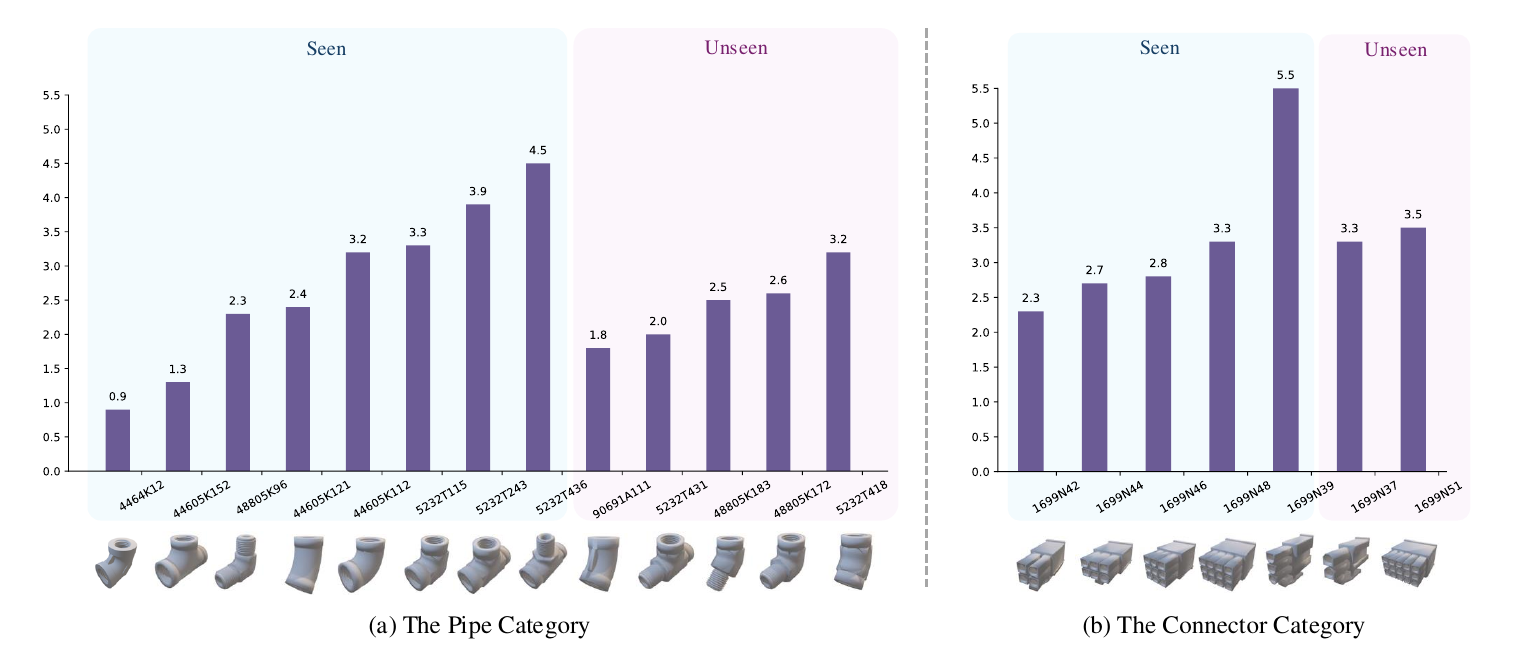}
\end{center}
\caption{\textbf{Category-level sim-to-real evaluation accuracy.} For the pipe class, we train on the first \rebuttal{8} objects and evaluate all 13 objects. For the connector class, we train on the first 5 objects and evaluate all 7 objects. 
\rebuttal{We report ADD-S (mm) and ADD (mm) errors for symmetric and non-symmetric objects, respectively. Lower ADD/ADD-S error implies better performance.}
}
\label{fig:category_acc}
\vspace{-15pt}
\end{figure*}

\vspace{-5pt}
\subsection{Generalizable Tactile Pose Estimation to Unseen Intra-category Objects}
\vspace{-5pt}

Having validated our method for instance-level pose estimation, we next assess its ability to generalize to unseen objects after training on multiple objects from a single category.
To this end, we conduct a proof-of-concept experiment on two categories: \textit{Pipe} and \textit{Connector}, training on 8 and 5 objects, respectively, and testing on 5 and 2 objects. Despite training on a small dataset, our energy network learns generalizable representations due to shared local features within each category. As shown in Fig.~\ref{fig:category_acc}, performance on unseen instances does not degrade significantly, suggesting that our method can generalize to new instances within a category and has the potential for broader generalization with large-scale training.

\vspace{-10pt}
\subsection{Validation of Uncertainty Estimation, Pose Tracking, and Arbitrary Contacts}
\vspace{-10pt}

We also validate the capabilities of our framework in uncertainty estimation, pose tracking, and handling arbitrary contact scenarios. Our experiments demonstrate that the framework effectively supports in-hand pose tracking with minimal error and stable frame rates. Additionally, our uncertainty estimation method consistently outperforms baseline approaches, and the system can generalize well to situations involving arbitrary subsets of tactile observations.
Due to the page limit, we defer the detailed experimental results and additional discussions to the Appendix~\ref{appendix:addtional_exps}.

\vspace{-10pt}
\section{Conclusion} 
\vspace{-10pt}
\label{sec:conclusion}
In this work, we investigate tactile in-hand object pose estimation, a crucial task for high-precision pick-and-place manipulation.  
While existing approaches predominantly rely on regression, feature matching, or registration techniques, achieving high accuracy while maintaining adaptability to unseen CAD models remains a significant challenge.  
We propose a novel three-stage framework: the first stage involves sampling and pre-ranking pose candidates, followed by iterative refinement of these candidates in the second stage. In the final stage, post-ranking is applied to determine the most probable pose.  
All stages are governed by a unified energy-based diffusion model integrated with a render-compare architectural design, trained solely on simulated data. This approach alleviates the laborious and expensive process of real-world data collection.
Extensive experimental evaluations show that our method surpasses conventional regression, matching, and registration-based baselines while demonstrating strong generalizability to previously unseen intra-category CAD models. Ablation studies further confirm that the render-compare design significantly enhances sim-to-real performance.  
Additionally, we demonstrate that our framework extends naturally to pose tracking, uncertainty estimation, and arbitrary tactile input.  
To facilitate future research, we will open-source our training and inference pipeline, along with training datasets and real-world evaluation data.

\clearpage

\textbf{Limitations and Future Work.}
Our framework has two main limitations.
First, although our pose tracking algorithm operates at 10 FPS during inference, the pose estimation procedure runs at less than 1 FPS, taking 1 to 2 seconds per pose estimate. This is significantly slower compared to methods such as~\cite{bauza2023tac2pose}. This limitation could be alleviated by employing more advanced diffusion models, such as Flow-Matching~\cite{lipman2022flow}. While our approach requires over 500 steps of denoising during inference, Flow-Matching only requires tens of steps, making it several orders of magnitude faster. This suggests that it may be possible to accelerate our pipeline to over 25 FPS in the future.
Second, our experiments have been conducted at the category level on a small set of objects from a single category. Future work will focus on validating our approach on a larger set of objects and assessing its generalization to novel objects from unseen categories.


\acknowledgments{
We thank Jiyao Zhang, Tianhao Wu, and Jinzhou Li for their insightful discussions.
This project was supported by the National Youth Talent Support Program (8200800081) and the National Natural Science Foundation of China (62376006).
}


\bibliography{references}

\clearpage

\begin{appendix}

\appendix

\section{Synthesizing Tactile Pose Estimation Dataset}
\label{appendix:sim_dataset}
\begin{wrapfigure}{tr}{0.60\textwidth}
\begin{center}
\vspace{-20pt}
\includegraphics[width=\linewidth, trim=0cm 0cm 0cm 0cm, clip]{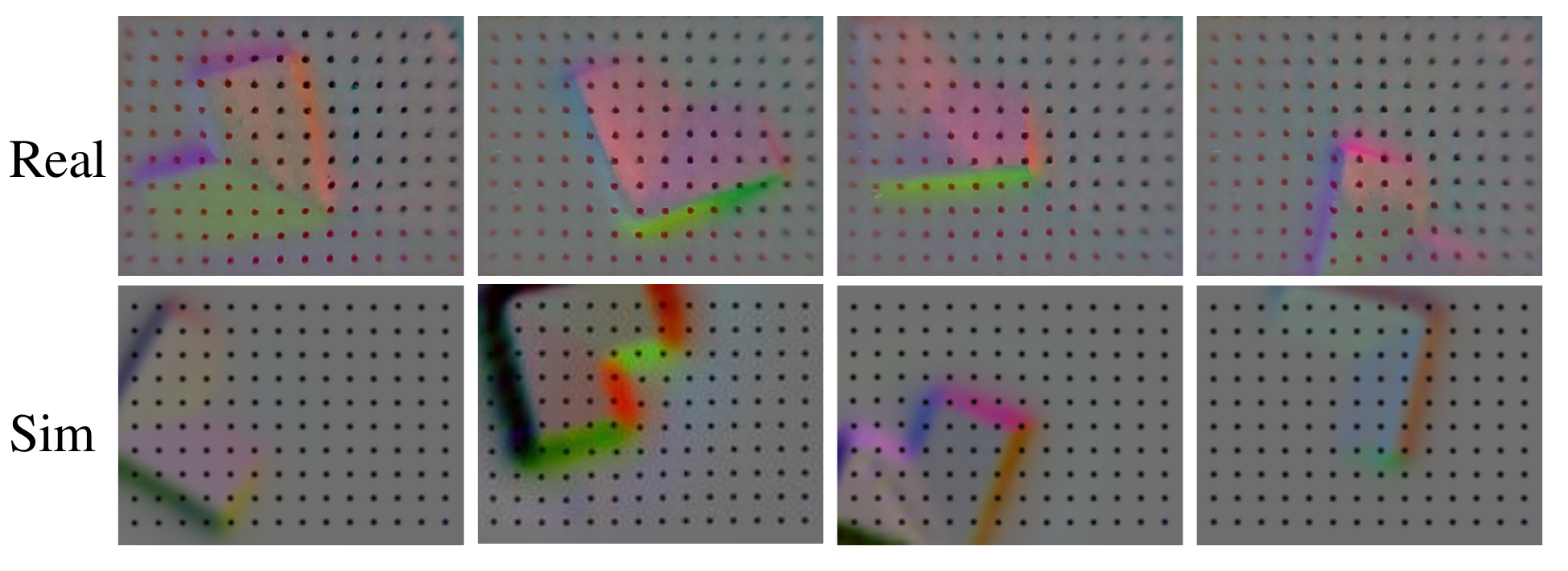}
\end{center}
\vspace{-10pt}
\caption{
Simulation and real-world tactile images.
}
\label{fig:sim_real_obs}
\vspace{-15pt}
\end{wrapfigure}
We generate a tactile pose estimation dataset under practical conditions, where the objects are securely grasped in hand.
As illustrated in Fig.~\ref{fig:pipeline} (I), for each object $O$, we first select several grasping approaches based on the object's canonical frame, such as $X+$ or $Y-$, which indicate that the left sensor will face the object along the corresponding axis for grasping. 
The object is initially positioned at the center of the TCP (Tool Center Point) frame. To introduce variability, randomization is applied in two dimensions: (1) along the $xy$-plane perpendicular to the grasping approach direction, and (2) in the rotational direction around the grasping approach axis. 
Specifically, the position in the $xy$-plane is randomized within $[ -b_m, b_m ]$, where $b_m$ represents the maximum bounding box of the object in the given $xy$-plane. 
The gripper subsequently approaches the object along the selected grasping direction, with an indentation randomized within $[0.2\mathrm{mm}, 1\mathrm{mm}]$. 
Once the grasp is performed, we render the FEM-simulated left and right RGB tactile images. 
Finally, we filter out data points where the contact region is smaller than 5\% of the sensor's imaging area to ensure data quality.
\rebuttal{Notably, we observed that ColorJitter serves as an effective augmentation strategy during training across all methods presented in this paper, as demonstrated by the ablation studies in Sec.~\ref{sec:baseline-ablations}. Specifically, we utilized the PyTorch implementation of ColorJitter with brightness, contrast, saturation, and hue parameters set to 0.3, 0.3, 0.3, and 0.1, respectively.}


\section{Real-World Data Collection and Hardware Setup}
\label{appendix:real_world_setup}

\begin{wrapfigure}{tr}{0.30\textwidth}
\begin{center}
\vspace{-20pt}
\includegraphics[width=\linewidth, trim=0cm 0cm 0cm 0cm, clip]{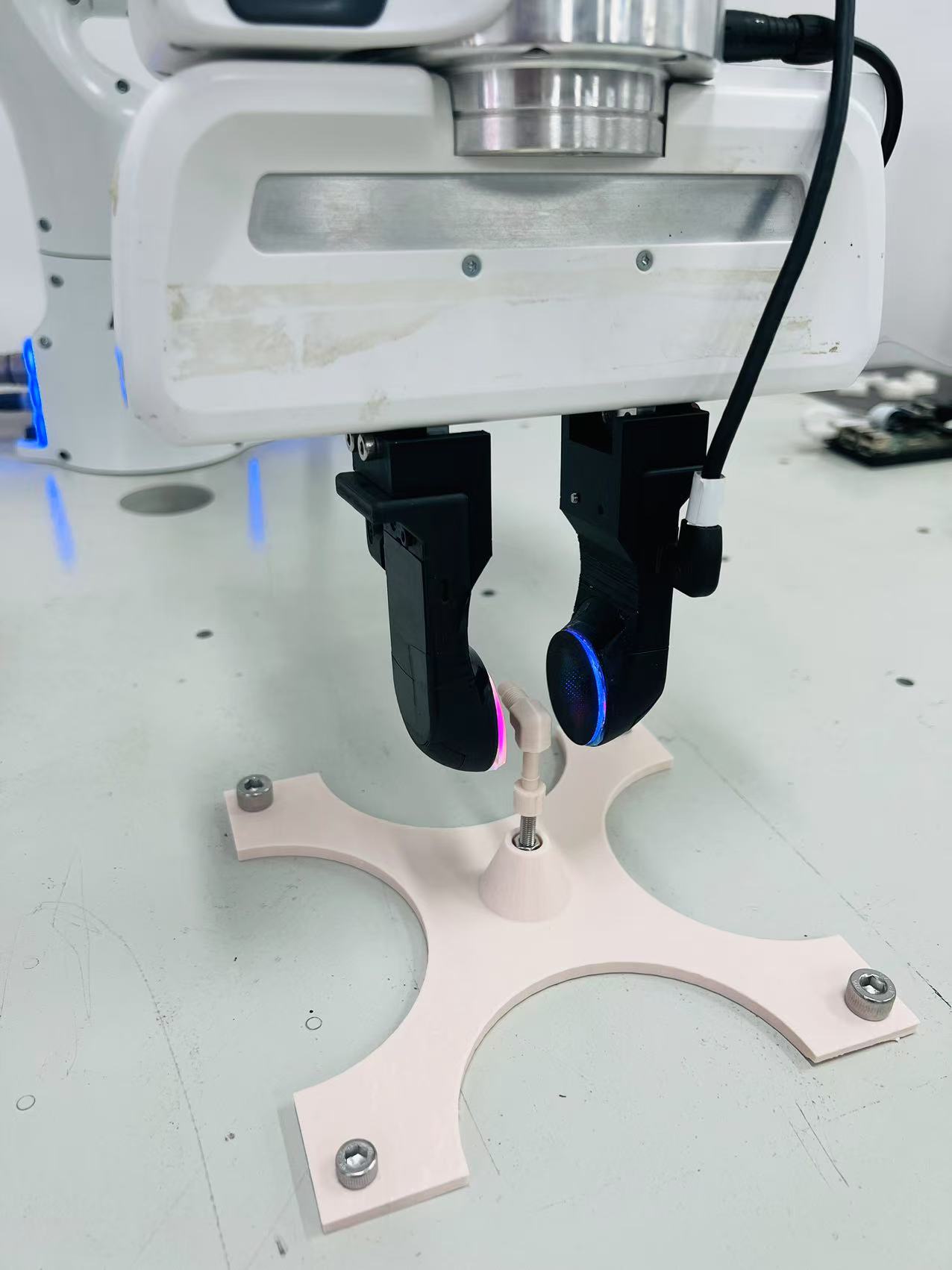}
\end{center}
\vspace{-10pt}
\caption{
Hardware setup.
}
\label{fig:hardware}
\vspace{-20pt}
\end{wrapfigure}
\textbf{Hardware Setup.}
We use the GelSlim 3.0 sensor for both simulation and real-world data collection, following the approach of~\cite{bauza2023tac2pose}. This sensor provides high-resolution tactile readings in the form of RGB images. It consists of a deformable membrane that responds to contact, and an internal camera that captures the deformation. The sensor transmits tactile observations via ROS as $640 \times 480$ compressed images at a frequency of 90Hz.
The object CAD model is 3D-printed and securely mounted on a table using a fixture. The sensor is attached to the end-effector of a Franka Panda robot, which randomly samples poses to establish contact with the fixed object model.

\textbf{Real-world Data Collection.} We collected labeled datasets of tactile observations for grasps performed on 30 objects. Each object was mounted in a known position and orientation. For each dataset, we collected pairs of RGB tactile images and their corresponding ground-truth object poses relative to the gripper center. A comparison of tactile imprints obtained from simulation and real-world experiments is shown in Fig.~\ref{fig:sim_real_obs}.
The robotic system used for data collection includes the Franka Panda robot equipped with a GelSlim 3.0 tactile sensor on each finger. The hardware setup is illustrated in Fig.~\ref{fig:hardware}.
\textbf{Data Collection Process.}
The data collection procedure consists of the following steps:

\begin{itemize} 
\item \textbf{Mounting the Object:}
Each object is attached to a fixed platform using its threaded base. This ensures that the object remains stationary during grasping operations and prevents any slippage.

\item \textbf{Calibrating Object Pose:}  
We initially replicate the real-world experimental setup within a simulation environment to estimate the object’s pose relative to the gripper center. The pose is then refined by performing multiple grasps at different orientations and indentation depths. We iteratively compare tactile images from real-world grasps with those from the simulation and adjust the simulated object pose until both sets of images align closely.

\item \textbf{Data Collection:}  
For each object, predefined grasp axes and approach directions are established. We systematically gather tactile data by performing grasps along these directions, varying the gripper’s x, y, and $\theta$ positions relative to the object. To simulate real-world uncertainties, we introduce in-plane rotational noise in each grasp. We also conduct grasps at three distinct indentation depths to capture variations in grasp quality and applied force.
\end{itemize}

\section{Baseline Implementation}
\label{appendix:baseline}
\textit{FilterReg} utilizes a partial point cloud extracted from the depth image produced by the tactile sensor. It is a probabilistic variant of the traditional ICP algorithm, aligning the tactile point cloud with the CAD model within a Gaussian Mixture Model (GMM) framework. FilterReg iteratively minimizes alignment errors to refine the estimated pose. For this method, we provide a rough estimate by introducing noise to the ground truth pose. \rebuttal{We implement two variants of FilteReg: FilteReg(Global) matches the tactile point cloud with the complete point cloud derived from the CAD model. FilteReg(Partial) uses the oracle object pose and grasping depth to acquire partial point cloud from the CAD model, serving as an upper bound for FilterReg.}

\textit{Vanilla Regression} is a simple baseline where the pose is regressed directly from the tactile RGB image and the point cloud derived from the CAD model. The point cloud is encoded using PointNet, and the tactile image is processed through a combination of convolutional and Vision Transformer blocks, similar to our approach. Features from both modalities are fused and passed through a regression network to predict the pose.

\textit{Matching} \rebuttal{We reimplement Tac2Pose~\cite{bauza2023tac2pose}, as the original implementation has not been publicly released. For each object, we construct pose grids with a resolution of 2.5 mm and 6 degrees, resulting in approximately 8K to 17K grid points depending on the object geometry. The MoCo module~\cite{he2020momentum} is trained for 30 epochs, ensuring convergence of the training loss for all objects. For fair comparison, the main differences from the original method are: (1) we omit training the image-to-image transfer module, thus making our method purely simulation-based; and (2) during real-world inference, we directly utilize the raw contact masks derived from depth measurements provided by the tactile sensor.}

\section{Extension for Pose Tracking and Uncertainty Estimation} 
\label{appendix:extension}

In this section, we provide a detailed description of the extensions to our framework for supporting pose tracking and uncertainty estimation. These additions are crucial for improving the robustness of object manipulation, particularly when dealing with dynamic environments and initial contact uncertainty.

\subsection{Pose Tracking} In precise manipulation tasks, objects are often subject to motion, and it is essential to track their poses continuously over time to avoid unintended collisions with the environment. Our pose estimation framework can be naturally extended to pose tracking by modifying the prior distribution during the pre-filtering stage.

\textbf{Tracking Prior Distribution.} In the pose tracking scenario, the primary challenge is to incorporate temporal continuity from previously estimated poses. The tracking prior distribution is designed to reflect this temporal consistency, making it more informative compared to the prior used for the initial pose estimation.

Given the previously estimated in-hand pose $\cand_t$, we define the tracking prior distribution for the next frame as a Gaussian distribution:

\begin{equation} \begin{aligned} \trackprior(\initcand_{t+1}) = \mathcal{N}(\initcand_{t+1}; \cand_{t}, \sigma_{track}^2 \mathbf{I}) \end{aligned} \label{eq:tracking_prior} \end{equation}

Here, $\sigma_{track} = 0.05$ is a hyperparameter that controls the spread of the distribution. This prior encourages the next pose to remain close to the previous estimate while allowing for some flexibility to account for small changes in pose due to motion or noise in the sensing.

\textbf{Pose Tracking Procedure.} Upon receiving a tactile observation $T_{t+1}$ at frame $t+1$, we initialize the particle filter-based ODE solver (PF-ODE) in Eq.~\ref{eq:reverse_sde} with $K$ candidates sampled from the tracking prior $\trackprior$. This approach leverages the previous pose estimate to more effectively predict the current pose.

Since pose tracking assumes continuity between consecutive frames, the tracking prior is typically closer to the high-density regions of the posterior distribution $\dist(\pose | \obj, T)$, compared to the prior used for initial pose estimation. This allows the pose tracking process to be more efficient.

To further enhance the tracking efficiency, we set a smaller value for the time step parameter $t_0$ in the PF-ODE, specifically $t_0 = 0.1$, during tracking. This smaller time step allows the model to utilize more informative gradients, which leads to faster convergence during tracking compared to initial pose estimation.

\textbf{Pose Selection.} At the end of the tracking process, we select the pose candidate with the highest energy value as the final estimate. This candidate is the one that best matches the observed tactile feedback, considering both the prior distribution and the sensory information.

\textbf{Efficiency of Pose Tracking.} Compared to frame-by-frame pose estimation, pose tracking is significantly faster, with an inference rate of approximately 10 Hz, which is about 10 times faster than the initial pose estimation pipeline. This speed improvement comes from the reduced number of refinement steps required during tracking, as the previously estimated pose serves as a strong prior.

In practice, it is most beneficial to spend additional time accurately estimating the initial pose at the first contact frame, where the uncertainty is typically higher. Once the initial pose is accurately determined, subsequent frames can be processed efficiently with minimal refinement, allowing for real-time tracking.

\subsection{Uncertainty Estimation} Uncertainty plays a crucial role in object manipulation, particularly when the object is grasped at uncertain contact points. Our framework quantifies uncertainty by calculating the variance of the refined pose candidates. The goal is to identify regions with lower uncertainty and guide the robot to re-grasp the object at those locations to improve pose estimation accuracy.

\textbf{Relative Uncertainty Estimation.} The uncertainty of the pose candidates $\{\cand_i\}^K_{i=1}$ is measured by computing their variance. This is done by first aggregating the candidate poses into a mean pose, denoted as $\cand_{\text{mean}}$, following the procedure described in GenPose~\cite{zhang2024generative}.

Once the mean pose is computed, we define the uncertainty as the variance of the pose candidates. Specifically, the uncertainty $S^2$ is calculated as:

\begin{equation} S^2 = \frac{1}{K} \sum_{i=1}^{K} d(\cand_i, \cand_{\text{mean}}, \obj) \label{eq:uncertainty} \end{equation}

Here, $d(\cand_i, \cand_{\text{mean}}, \obj)$ is a distance metric that quantifies the difference between the candidate pose $\cand_i$ and the mean pose $\cand_{\text{mean}}$ with respect to the object $\obj$. For non-symmetric objects, we use the Average Distance Difference (ADD) metric, while for symmetric objects, we use the ADD-S metric. Both of these metrics measure the average Euclidean distance between the model points under two different poses.

\textbf{Grasp Comparison and Re-grasping.} Given $N$ different grasp candidates ${\g_i}{i=1}^N$ that result in different tactile images ${T_i}{i=1}^N$ of the object $\obj$, we compare the relative uncertainty of each grasp based on the computed variance $S^2_i$ of their corresponding pose candidates.

The relative uncertainty between two grasps $\g_i$ and $\g_j$ is determined as:

\begin{equation} \g_i \succ \g_j \iff S^2_i > S^2_j, \quad \g_{re} = \argmin_{i} S^2_i \end{equation}

In this context, the grasp with the lowest uncertainty, $\g_{re}$, is selected for re-grasping. This means that the robot will choose the grasp pose that minimizes uncertainty in the pose estimation, leading to more accurate and reliable manipulation.

\textbf{Re-grasping for Improved Pose Estimation.} The ability to re-grasp the object at regions of lower uncertainty is critical for improving pose estimation accuracy over time. This approach allows the robot to refine its understanding of the object’s pose by re-grasping at more stable and distinguishable contact regions, reducing the effect of initial uncertainty and improving the overall robustness of the manipulation process.

\section{Additional Experimental Details}
\label{appendix:addtional_exps}

In this section, we provide detailed results and discussions for the experiments related to in-hand pose tracking, uncertainty estimation, and the handling of arbitrary contacts.

\subsection{Pose Tracking and Uncertainty Estimation}

We conduct extensive studies to evaluate the effectiveness of our method in two key aspects: in-hand pose tracking and relative uncertainty estimation.

\textbf{In-hand Pose Tracking.}
We collect three real-world trajectories on three objects, each consisting of 50 time steps. The initial pose estimate for each tracking trajectory is obtained using our pose estimation method, and subsequent estimations are computed as described in Appendix~\ref {appendix:extension}. As shown in Tab.~\ref{table:tracking}, our method demonstrates robust performance, maintaining an error within 2mm across all three objects. Additionally, we observe a stable pose tracking frame rate of 10 Hz.

\ifdefined\isMainDocument
\else
\documentclass{article}

\usepackage{array}
\usepackage{multicol}
\usepackage{multirow}
\usepackage{graphicx}
\usepackage{makecell}

\newcommand\mydata[2]{$#1_{\pm#2}$}
\begin{document}

\fi

\begin{table}[h]
\centering
\renewcommand{\arraystretch}{1.5} 
\begin{tabular}{|c|c c c|}
  \hline
  & Bear Housing(mm) & Rail(mm) & Deutsch Connector(mm) \\
  \hline
  Ours & 1.2 & 1.8 & 1.5 \\
  \hline
\end{tabular}

\caption{Pose Tracking Results.}
\label{table:tracking}
\end{table}

\ifdefined\isMainDocument
\else
\end{document}
\fi

\textbf{Relative Uncertainty Estimation.}
We sample 10 sets of data points from the real-world test set of three objects, with each set containing data points from 10 different grasps. For each set, we apply the relative uncertainty estimation method (described in Appendix~\ref {appendix:extension}) to select the Top-1, Top-3, and Top-5 grasps and compute the average pose error for each selection. As shown in Tab.~\ref{table:confidence}, the pose errors for grasps selected using uncertainty estimation consistently outperform the baseline, where a grasp is randomly chosen from the 10 candidates. Furthermore, as the Top-K selection narrows, the average error decreases, demonstrating the effectiveness of the uncertainty estimation method.
\ifdefined\isMainDocument
\else
\documentclass{article}

\usepackage{array}
\usepackage{multicol}
\usepackage{multirow}
\usepackage{graphicx}
\usepackage{makecell}

\newcommand\mydata[2]{$#1_{\pm#2}$}
\begin{document}

\fi

\begin{table}[h!]
\centering
\renewcommand{\arraystretch}{1.5} 
\begin{tabular}{|c|c c c|}
  \hline
  & Nut (mm) & Cotter (mm) & Cable Clip (mm) \\
  \hline
  Random Selection & 2.8 & 2.9 & 9.9 \\
  Top-1 Confidence & 1.5 & 0.6 & 1.5 \\
  Top-3 Confidence & 2.1 & 0.8 & 4.7 \\
  Top-5 Confidence & 2.6 & 1.0 & 7.7 \\
  \hline
\end{tabular}

\caption{Grasp uncertainty estimation results. We compute the variance of estimated poses of a certain grasp generated by our model. Lower variance indicates lower uncertainty (higher confidence). We compute the mean ADD-S(mm) over top-k confident grasps. The ADD-S error of top-k grasps are lower than the mean error of random selected grasp, demonstrating the effectiveness of our grasp uncertainty estimation.}
\label{table:confidence}
\vspace{-20pt}
\end{table}

\ifdefined\isMainDocument
\else
\end{document}
\fi

\subsection{Extending UniTac2Pose to Arbitrary Contacts}
Our framework is designed to handle an arbitrary number of tactile contacts, thanks to its end-to-end training paradigm. To validate this, we compare three variants of the framework across six objects:

\begin{itemize} \item \textit{Double}: The original version of the framework, which uses two tactile images as input. \item \textit{Single}: A variant that ablates the right sensor observation, using only the left tactile sensor for render-compare. \item \textit{Arbitrary}: A variant where either the left or right tactile image is randomly masked during training and testing. \end{itemize}

As shown in Tab.~\ref{table:arbitraty_contact}, \textit{Double} consistently outperforms both \textit{Single} and \textit{Arbitrary}, as the latter two suffer from higher observation ambiguity. Although \textit{Arbitrary} performs poorly on the \textit{Stud} object, its overall performance is comparable to \textit{Single}, demonstrating that our method can generalize well to scenarios where only a subset of tactile observations is available.
\ifdefined\isMainDocument
\else
\documentclass{article}

\usepackage{array}
\usepackage{multicol}
\usepackage{multirow}
\usepackage{graphicx}
\usepackage{makecell}

\newcommand\mydata[2]{$#1_{\pm#2}$}
\begin{document}

\fi

\begin{table}[h!]
\centering
\renewcommand{\arraystretch}{1.5} 
\begin{tabular}{|c|c c c|}
    \hline
    &  Single~(mm)  & Double~(mm) & Arbitrary~(mm)\\\hline 
    Cotter & 1.8 & 1.5 & 2.5    \\ \hline    

    Hose & 2.9 & 1.2 & 2.4    \\ \hline

    Hydraulic & 2.3 & 2.4 & 2.5    \\ \hline

    Round Nut & 2.9 & 2.4 & 3.0    \\ \hline

    Rail & 1.3 & 1.5 & 1.2    \\ \hline

    Stud & 1.6 & 2.2 & 8.6    \\ \hline
\end{tabular}
\caption{Evaluation with different contact settings.}
\label{table:arbitraty_contact}
\end{table}







\ifdefined\isMainDocument
\else
\end{document}
\fi





\section{Ablations on three stages.}
\label{appendix:ablation_three_stages}
We conduct ablation studies on the three stages of our method: pre-filtering, refinement, and post-ranking. We also compare refining only the top candidate pose. As shown in Tab.~\ref{table:three_stage}, our full method surpasses all variants, demonstrating the effectiveness of each stage.

\def\isMainDocument{}
\ifdefined\isMainDocument
\else
\documentclass{article}

\usepackage{array}
\usepackage{multicol}
\usepackage{multirow}
\usepackage{graphicx}
\usepackage{makecell}
\usepackage{amssymb}

\newcommand\mydata[2]{$#1_{\pm#2}$}
\begin{document}

\fi

\begin{table}[h!]
\centering
\renewcommand{\arraystretch}{1.5} 
\scriptsize
\vspace{-5pt}
\setlength{\tabcolsep}{4pt}
\begin{tabular}{|c|c| c c c c|}
  \hline
  &  Ours & w/o pre-filter & w/o refine & w/o post-rank & refine top-1\\
  \hline
  Bear Housing & \textbf{2.4} & \underline{2.8} & 13.1 & 3.1 & 3.7 \\
  Cable Holder & \textbf{2.4} & \underline{2.7} & 12.0 & 3.3 & 4.2 \\
  Hose & \textbf{1.5} & \underline{1.6} & 16.0 & 1.7 & \underline{1.6} \\
  \hline
\end{tabular}
\caption{\footnotesize Ablation on three stages. Ours \textit{w/o pre-filter} randomly samples 16 candidates from the prior, \textit{w/o refine} selects top-1 candidate as the output, \textit{w/o post-rank} reports the average ADD-S of 16 refined poses, and \textit{refine top-1} means refining the top-1 candidate as the output.} 
\label{table:three_stage}
\end{table}

\ifdefined\isMainDocument
\else
\end{document}
\fi

\section{Choices of t.} 
\label{appendix:ablation_t}

\begin{figure}[h]
\begin{center}
\begin{minipage}{0.49\linewidth}
    \centering
    \includegraphics[width=1\linewidth, trim=0cm 0cm 0cm 0cm, clip]{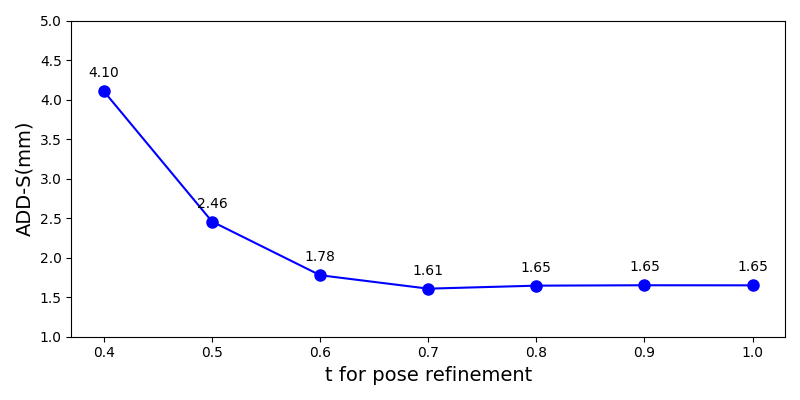}
    \vspace{-10pt}
    \caption{Evaluation of different initial t for pose refinement.}
    \label{fig:t_refine}
\end{minipage}
\hfill
\begin{minipage}{0.49\linewidth}
    \centering
    \includegraphics[width=1\linewidth, trim=0cm 0cm 0cm 0cm, clip]{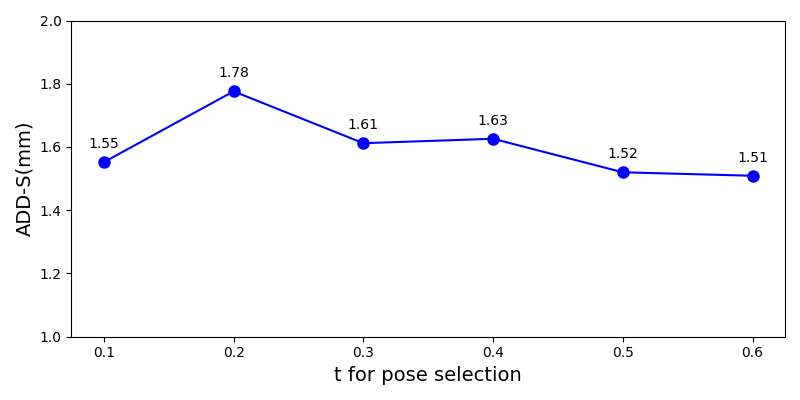}
    \vspace{-10pt}
    \caption{Evaluation of different t for pose selection.}
    \label{fig:t_selection}
\end{minipage}
\end{center}
\vspace{-10pt}
\end{figure}

For pose refinement, we train a model on the Hose object with the time parameter $t$ uniformly sampled from the range $[0, 1]$. During inference, we initialize the RK45 solver with $t$ values ranging from 0.4 to 1.0 in increments of 0.1 and evaluate the corresponding performance. As shown in Fig.~\ref{fig:t_refine}, initialization with $t \geq 0.6$ results in consistent performance, whereas $t \leq 0.5$ leads to a notable drop in accuracy. Based on this observation and inference efficiency, we choose $t = 0.6$ as the default setting.
For pose selection, we vary $t$ from 0.1 to 0.6 and evaluate performance across these values. As illustrated in Fig.~\ref{fig:t_selection}, our pose selection method demonstrates robustness to the choice of $t$.

\vspace{-10pt}
\section{Training UniTac2Pose on objects from multiple categories.} 
\vspace{-10pt}

We train a single model using data from three objects: Round Nut, Hose, and Rail. These objects were selected for their distinct shapes and differing symmetry properties—Round Nut exhibits half-turn symmetry, Rail has quarter-turn symmetry, and Hose is asymmetric. As shown in Tab.~\ref{table:overfit_3obj}, the multi-object model successfully fits all three objects, achieving performance comparable to individual models trained separately for each object. This result suggests that our method is capable of learning across multiple object categories within a single model and holds promise for improved out-of-distribution (OOD) generalization as more diverse training data becomes available.

\def\isMainDocument{}
\ifdefined\isMainDocument
\else
\documentclass{article}

\usepackage{array}
\usepackage{multicol}
\usepackage{multirow}
\usepackage{graphicx}
\usepackage{makecell}

\newcommand\mydata[2]{$#1_{\pm#2}$}
\begin{document}

\fi

\begin{table}[h!]
\centering
\renewcommand{\arraystretch}{1.5} 
\setlength{\tabcolsep}{6pt}
\begin{tabular}{|c|c c c|}

    \hline
    & Round Nut(mm) & Hose(mm) & Rail(mm) \\
    \hline
    Multi-object Model &  2.3 & 1.5 & 1.7 \\
    Single-object Model & 2.4 & 1.5 & 1.5 \\
    \hline
\end{tabular}
\caption{Comparison of a model trained on three objects versus models trained on individual objects.}
\label{table:overfit_3obj}
\end{table}

\ifdefined\isMainDocument
\else
\end{document}
\fi




\section{The sim-to-real performance gap.} 

We report the evaluation performance in both simulation and the real world in Tab.~\ref{table:sim-to-real}. For most objects, the models achieve comparable performance in simulation. However, the Hook object is an exception, as its uneven surface results in severe partial observations and ambiguities in the contact images, ultimately degrading performance. The sim-to-real performance gap varies across objects, largely due to differences in contact surface geometry, object size, and symmetry. Overall, our method demonstrates a relatively small sim-to-real gap, owing to the use of extensive data augmentation and randomization during training.

\def\isMainDocument{}

\ifdefined\isMainDocument
\else
\documentclass{article}

\usepackage{array}
\usepackage{multicol}
\usepackage{multirow}
\usepackage{graphicx}
\usepackage{makecell}

\newcommand\mydata[2]{$#1_{\pm#2}$}
\begin{document}

\fi

\begin{table*}[h!]
\centering
\renewcommand{\arraystretch}{1.5} 
\resizebox{\textwidth}{!}{
\begin{tabular}{|c|c c c c c c c c c c|}

    \hline
    & Bear Housing & Cable Holder&Cable Clip&Round Nut&Cotter&Hook&Hose&Hydraulic&Stud&Rail \\
    \hline
    Sim & 1.4 &  0.8 &  1.0 &  1.1 &  1.3 & 3.5 & 0.9 & 1.3 & 1.3 & 1.4 \\
    Real & 2.5 & 2.7 & 1.6 & 2.4 & 1.5 & 3.0 & 1.5 & 2.4 & 2.2 & 1.5 \\
    \hline
    $\Delta$ & 1.1 & 1.9 & 0.6 & 1.3 & 0.2 & -0.5 & 0.6 & 1.1 & 0.9 & 0.1\\
    \hline
\end{tabular}

}
\caption{The sim-to-real performance gap.We report ADD-S (mm) and ADD (mm) errors for symmetric and non-symmetric objects respectively. Lower ADD/ADD-S error implies better performance.}
\vspace{-15pt}
\label{table:sim-to-real}
\end{table*}

\ifdefined\isMainDocument
\else
\end{document}
\fi




\end{appendix}

\end{document}